\documentclass[journal,twoside,web]{ieeecolor}
\usepackage{generic}
\usepackage{cite}
\usepackage{amsmath,amssymb,amsfonts}
\usepackage{algorithmic}
\usepackage{graphicx}
\usepackage{textcomp}
\usepackage{comment}
\usepackage{hyperref}
\usepackage{url}
\usepackage{array}
\usepackage{epstopdf}
\usepackage{soul}
\soulregister\cite7
\soulregister\ref7
\soulregister\pageref7
\usepackage[table]{xcolor}
\usepackage[geometry]{ifsym}
\let\labelindent\relax
\usepackage{enumitem}
\usepackage{dblfloatfix}
\usepackage{bm}
\usepackage[flushleft]{threeparttable}
\usepackage{multirow}
\usepackage{capt-of}
\usepackage{url}
\def\BibTeX{{\rm B\kern-.05em{\sc i\kern-.025em b}\kern-.08em
    T\kern-.1667em\lower.7ex\hbox{E}\kern-.125emX}}
\begin{document}
\title{Relative Afferent Pupillary Defect Screening through Transfer Learning}
\author{Dogancan Temel, \IEEEmembership{Member, IEEE}, Melvin~J.~Mathew, \IEEEmembership{Member, IEEE}, Ghassan AlRegib, \IEEEmembership{Senior Member, IEEE}, and Yousuf M. Khalifa
\thanks{D. Temel, M. J. Mathew and G. AlRegib are with the School of Electrical and Computer Engineering, Georgia Institute of Technology, Atlanta, GA, 30332-0250 USA. Y. M. Khalifa is with the Emory University School of Medicine and Grady Memorial Hospital. Corresponding author email: cantemel@gatech.edu.This work was supported by Georgia Research Alliance.}}

\twocolumn[{%
\vspace{40mm}
{ \large
\begin{itemize}[leftmargin=2.5cm, align=parleft, labelsep=2.0cm, itemsep=4ex,]

\item[\textbf{Citation}]{D. Temel, M. J. Mathew, G. AlRegib and Y. M. Khalifa, "Relative Afferent Pupillary Defect Screening
through Transfer Learning," IEEE Journal of Biomedical and Health Informatics, 2019.}

\item[\textbf{DOI}]{https://doi.org/10.1109/JBHI.2019.2933773}

\item[\textbf{Bib}]  {@INPROCEEDINGS\{Temel2019\_BHI,\\ 
author=\{D. Temel and M. J. Mathew and G. AlRegib and Y. M. Khalifa\},\\ 
booktitle=\{IEEE Journal of Biomedical and Health Informatics \},\\
title=\{Relative Afferent Pupillary Defect Screening
through Transfer Learning\},\\ 
year=\{2019\},\\ 
doi=\{10.1109/JBHI.2019.2933773\},\\ 
}

\item[\textbf{Copyright}]{\textcopyright 2019 IEEE. Personal use of this material is permitted. Permission from IEEE must be obtained for all other uses, in any current or future media, including reprinting/republishing this material for advertising or promotional purposes,
creating new collective works, for resale or redistribution to servers or lists, or reuse of any copyrighted component
of this work in other works. }

\item[\textbf{Contact}]{\href{mailto:alregib@gatech.edu}{alregib@gatech.edu}~~~~~~~\url{https://ghassanalregib.com/}\\ \href{mailto:dcantemel@gmail.com}{dcantemel@gmail.com}~~~~~~~\url{http://cantemel.com/}}
\end{itemize}
\thispagestyle{empty}
\newpage
\clearpage
\setcounter{page}{1}
}
}]

\twocolumn[{%
\renewcommand\twocolumn[1][]{#1}%

\newcommand\blfootnote[1]{%
  \begingroup
  \renewcommand\thefootnote{}\footnote{#1}%
  \addtocounter{footnote}{-1}%
  \endgroup
}

\maketitle

\vspace{-12mm}
\centering
\includegraphics[width=0.85\linewidth]{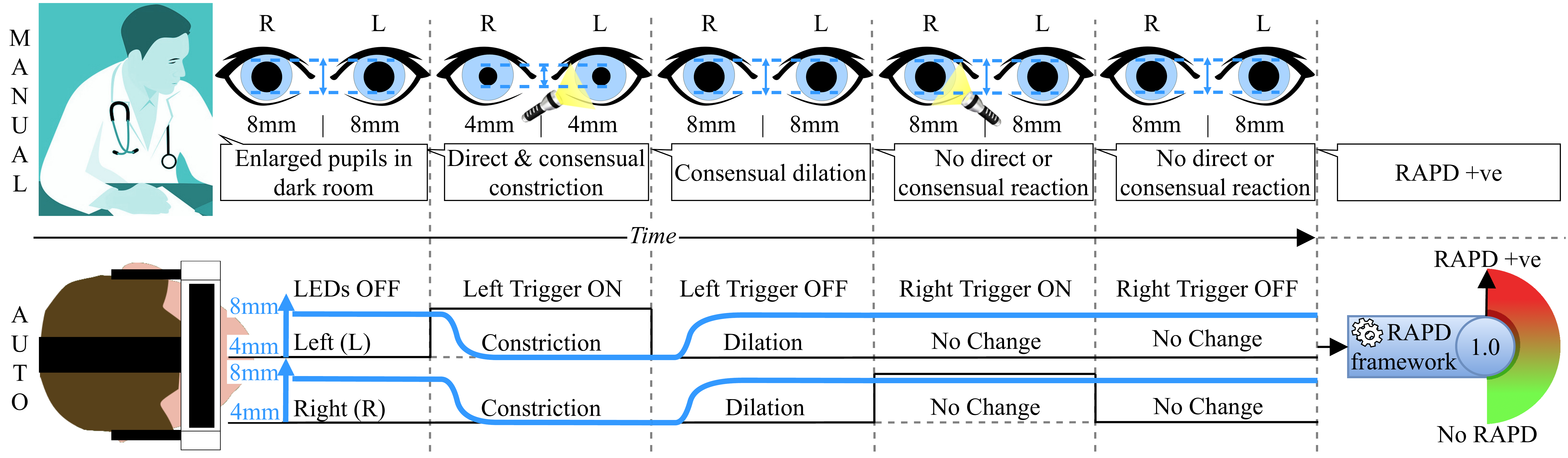}

\captionof{figure}{RAPD screening of a patient with positive RAPD on the right eye. Subjective examination is based on the visual assessment of the manual swinging flashlight test whereas objective examination is based on the automated swinging flashlight test combined with algorithmic analysis.}
\label{fig:rapd_high_level}
\vspace{2mm}
}]

\begin{abstract}
Abnormalities in pupillary light reflex can indicate optic nerve disorders that may lead to permanent visual loss if not diagnosed in an early stage. In this study, we focus on relative afferent pupillary defect (RAPD), which is based on the difference between the reactions of the eyes when they are exposed to light stimuli. Incumbent RAPD assessment methods are based on subjective practices that can lead to unreliable measurements. To eliminate subjectivity and obtain reliable measurements, we introduced an automated framework to detect RAPD. For validation, we conducted a clinical study with \texttt{lab-on-a-headset}, which can perform automated light reflex test. In addition to benchmarking handcrafted algorithms, we proposed a transfer learning-based approach that transformed a deep learning-based generic object recognition algorithm into a pupil detector. Based on the conducted experiments, proposed algorithm \texttt{RAPDNet} can achieve a sensitivity and a specificity of 90.6\% over 64 test cases in a balanced set, which corresponds to an AUC of 0.929 in ROC analysis. According to our benchmark with three handcrafted algorithms and nine performance metrics, \texttt{RAPDNet} outperforms all other algorithms in every performance category.

\end{abstract}


\section{Introduction}
\label{sec:intro}
\IEEEPARstart{A}{pproximately} 1.3 million Americans are considered blind, which is expected to increase to 2.2 million by 2030 according to the National Eye Institute statistics \cite{NEI2014}. Another 2.9 million Americans have low vision, which is estimated to be 5 million by 2030. The rapidly expanding geriatric population, the increased burden of eye diseases, and time spent in front of digital displays significantly increase the patient pool size \cite{NEI2014}. Based on the American Academy of Ophthalmology survey, there were more than 35 million eye disease cases in 2010 \cite{AAO2018} whereas the number of active ophthalmologists was less than 20 thousand in United States. Existing ocular technologies are not suitable for large-scale examinations worldwide because of their immobility, cost, lack of automation, and limited capabilities. In-person tests require experienced personnel and variation between subjective opinions can lead to inconsistencies and inaccuracies. Therefore, there is an emerging need for low-cost, portable, programmable, and multi-purpose vision monitoring and testing systems that conduct consistent and accurate eye exams.

In this study\footnote{This study is supported by the Georgia Research Alliance. The authors of this paper,  Georgia Institute of Technology and Emory University are entitled to royalties in case of commercialization of the developed technology \cite{Temel2018_RAPD_prov,Temel2019_RAPD_nonprov,Temel2019_transfer_prov}.}, we focus on pupillary light reflex assessment, which can be an indicator of numerous conditions including but not limited to optic nerve disorders, trauma, autism, alcohol and recreational drugs, exposure to toxins, and response to infections \cite{hall2018}. Specifically, we investigate the RAPD condition, which corresponds to relative afferent pupillary defect. RAPD is based on the differences between the reactions of the eyes when they are exposed to light \cite{Broadway2012}. In healthy subjects, light stimulation on one eye should lead to equal constriction of both pupils and constricted pupils should enlarge equally when there is no stimulation. In case of positive RAPD, patient's pupils constrict less or do not constrict when light stimuli swings from the unaffected eye to the affected eye as illustrated in Fig.~\ref{fig:rapd_high_level}.

Traditionally, RAPD is tested with a swinging flashlight test in which practitioner asks the patient to fixate on a distant target in a dark environment, swings the light source back and forth between eyes, and observes the size of the pupils and reactions \cite{Broadway2012}. An alternative to swinging flashlight test is the utilization of filters such as neutral density (NDF) \cite{Wilhelm2011}. A set of filters is placed between the light and the eyes in the NDF test, and the density of the filter that compensates for pupillary defects indicates the RAPD level. Even though existing methodologies are easy to practice, limited control over the environment and the subjectivity of the procedures can affect the reliability of the RAPD test significantly. A study conducted with experienced nurses in a postanesthesia care unit showed that pupillary examination findings significantly vary depending on the light source \cite{Omburo2017}. Moreover, practitioners have difficulty in detecting subtle abnormalities in case of small pupils, dark irises, and poorly reactive pupils \cite{Loewenfeld1971,Kawasaki1995}.

\begin{table*}[ht!]
\centering
\centering
\caption{Existing pupil datasets captured with closely-mounted setups and the RAPD-HD dataset.*}
\label{tab:datasets}
\begin{threeparttable}
\begin{tabular}{> {\centering\arraybackslash}p{1.6cm}>{\centering\arraybackslash}p{1.5cm}>{\centering\arraybackslash}p{1.6cm}>{\centering\arraybackslash}p{1.4cm}>{\centering\arraybackslash}p{1.4cm}>{\centering\arraybackslash}p{1.4cm}>{\centering\arraybackslash}p{1.4cm}>{\centering\arraybackslash}p{1.4cm}>{\centering\arraybackslash}p{2.2cm}>{\centering\arraybackslash}p{2.5cm}}
\hline
\textbf{}  & 
\textbf{\href{https://www.cl.cam.ac.uk/research/rainbow/projects/pupiltracking/datasets/}{Swirski}~\cite{Swirski2012}}  & \textbf{Kasneci }~\cite{Kasneci2014} & \textbf{Sippel~\cite{Sippel2014}}  & \textbf{\href{http://www.ti.uni-tuebingen.de/index.php?id=1827\&\&L=1}{ExCuSe}}~\cite{Fuhl2015_ExCuSe}   & \textbf{\href{http://www.ti.uni-tuebingen.de/index.php?id=1827\&\&L=1}{ElSe}}~\cite{Fuhl2016_ElSe} & \textbf{\href{http://www.ti.uni-tuebingen.de/index.php?id=1827\&\&L=1}{Microscope}~\cite{Fuhl2016_Microscope}}  &\textbf{~~~~~~~~LPW~\cite{Tonsen2016_LPW}} & \textbf{RAPD-HD}   \\ \hline

 \textbf{\begin{tabular}[c]{@{}c@{}}Number of\\ videos \end{tabular}} &-&40&20  &-&-&- &66 &64 \\ \hline

 %\textbf{FPS} &-&25&25&-&-&-&95 &28 \\ \hline

 \textbf{Resolution} &640x480&384x288&384x288&384x288&384x288&512x388&640x480&1,920x1,080\\ \hline

 \textbf{\begin{tabular}[c]{@{}c@{}}Number \\ of subjects\end{tabular}}    &2&40&20&17&7 & 2&22 & 24  \\ \hline

 \textbf{\begin{tabular}[c]{@{}c@{}}Acquisition \\ setup\end{tabular}}  &\begin{tabular}[c]{@{}c@{}}head-mounted\\ camera system\end{tabular}&\begin{tabular}[c]{@{}c@{}}head-mounted\\  eye tracker\end{tabular}&\begin{tabular}[c]{@{}c@{}}head-mounted\\  eye tracker\end{tabular}&\begin{tabular}[c]{@{}c@{}}head-mounted\\  eye tracker\end{tabular}&\begin{tabular}[c]{@{}c@{}}head-mounted\\  eye tracker\end{tabular}&microscope & \begin{tabular}[c]{@{}c@{}}head-mounted\\  eye tracker\end{tabular}&headset   \\ \hline

 \textbf{\begin{tabular}[c]{@{}c@{}}Subject \\ activity\end{tabular}}     &-&driving&shopping&\begin{tabular}[c]{@{}c@{}} driving (9)\\ shopping (8)  \end{tabular}&\begin{tabular}[c]{@{}c@{}} driving (5)\\ indoor (2)  \end{tabular}&-&\begin{tabular}[c]{@{}c@{}} staring at a\\ moving target  \end{tabular} &\begin{tabular}[c]{@{}c@{}} staring within\\ a headset  \end{tabular}    \\ \hline

 %\textbf{\begin{tabular}[c]{@{}c@{}}Subject \\ condition\end{tabular}} &-&\begin{tabular}[c]{@{}c@{}} Homonymous \\ visual field \\ defect (10),\\Glaucoma (10) \end{tabular}&\begin{tabular}[c]{@{}c@{}} binocular\\ Glaucoma (10)  \end{tabular}&-&-&-&\begin{tabular}[c]{@{}c@{}} visual \\ impairment (5)  \end{tabular}& \begin{tabular}[c]{@{}c@{}} Glaucoma (),\\ hyphema (),traumatic\\ mydriasis (),\\peripheral RPE\\ atrophy OU (),\\PVD OU  \end{tabular}   \\ \hline

  \textbf{Metadata} & \begin{tabular}[c]{@{}c@{}}pupil ellipse,  \\ center position\end{tabular}&\begin{tabular}[c]{@{}c@{}}center  \\ position\end{tabular}&\begin{tabular}[c]{@{}c@{}}center  \\ position\end{tabular}&\begin{tabular}[c]{@{}c@{}}center  \\ position\end{tabular}&\begin{tabular}[c]{@{}c@{}}center  \\ position\end{tabular}&\begin{tabular}[c]{@{}c@{}}center  \\ position\end{tabular} &\begin{tabular}[c]{@{}c@{}}pupil ellipse,  \\ center position\end{tabular} & \begin{tabular}[c]{@{}c@{}}swinging flashlight\\ and neutral density \\tests, medical history\end{tabular}    \\ \hline

\end{tabular}

\begin{tablenotes}
\item[*] Online sources of the datasets are hyperlinked to the dataset names in the first row of this table.
\end{tablenotes}

\end{threeparttable}
\vspace{-5mm}
\end{table*}

To eliminate shortcomings in terms of subjective bias and environmental conditions, digital pupilometers have been used in the literature \cite{Volpe2000,Miki2008,Waisbourd2015}. Even though these clinical studies shed a light on the utilization of automated methods, they do not disclose utilized algorithms or experimental data. Therefore, to design RAPD detection algorithms, it was necessary to develop a data acquisition platform and conduct a clinical study. Previously, we developed an RAPD detection algorithm based on Starburst \cite{Li2005_Starburst} algorithm  and validated its performance over 32 cases \cite{Temel2019_ISMR}. In this study, we extend the handcrafted algorithm benchmark with ExCuse \cite{Fuhl2015_ExCuSe} and ElSe \cite{Fuhl2016_ElSe} algorithms and propse a new algorithm that outperforms all the benchmarked algorithms. Overall, the contributions of this study are five folds.

\begin{itemize}[label=\textcolor{orange}{\FilledSmallSquare},leftmargin=*]

\item We provide a detailed comparison of existing pupil datasets in terms of setup, subjects, size, and metadata.

\item We benchmark the performance of three RAPD detection algorithms obtained from our framework using incumbent pupil localization and measurement techniques.

\item We introduce a transfer learning-based approach to convert generic visual recognition algorithms into pupil detection and localization algorithms.

\item We introduce an RAPD detection algorithm (RAPDNet) based on visual representations obtained from deep learning architecture AlexNet.  

\item Proposed algorithm RAPDNet achieves a sensitivity and a specificity of $90.6\%$ with an AUC of 0.929, which outperforms  all benchmarked algorithms over 64 test cases.

\end{itemize}

\vspace{1 mm}

\noindent{\bf Outline:} We briefly analyze existing pupil datasets along with related algorithmic approaches in Section~\ref{sec:related_work}. We describe the test dataset in Section~\ref{sec:rapd_dataset} and explain the RAPD detection framework in Section~\ref{sec:rapd_framework}. We discuss the experiments in Section~\ref{sec:experiments} and conclude our work in Section~\ref{sec:conc}.

\section{Related Work}
\label{sec:related_work}
The main mechanisms in an automated RAPD detection algorithm are can be divided into three main blocks as (i) pupil localization, (ii) pupil size measurement,  and (iii) pupil size comparison of left and right eyes in the test sequence. The majority of existing pupil localization algorithms require initialization of pupil location estimation based on fixed location \cite{Li2005_Starburst}, thresholding \cite{Fuhl2015_ExCuSe}, and pattern matching \cite{Swirski2012,Fuhl2016_ElSe}. After the initialization, preprocessing, edge detection, and shape fitting methods are commonly used to locate the pupils. Specifically, prepossessing leads to more descriptive representations, edge detection results in estimated pupil locations, and shape fitting eliminates false positive. Preprocessing techniques include adaptive thresholding \cite{Li2005_Starburst}, smoothing \cite{Li2005_Starburst,Fuhl2016_Microscope}, normalization \cite{Li2005_Starburst,Fuhl2016_ElSe}, morphological operations \cite{Swirski2012,Fuhl2015_ExCuSe,Fuhl2016_ElSe}.  After preprocessing stages, edge detection is usually applied over the entire image \cite{Li2005_Starburst}, region of interest (ROI) \cite{Swirski2012}, both ROI and entire image \cite{Fuhl2015_ExCuSe,Fuhl2016_ElSe}, and multiple-scales \cite{Fuhl2016_Microscope}. Pupil location estimations are obtained from edge maps and false positives are minimized with shape fitting methods including RANSAC-based ellipse fitting\cite{Li2005_Starburst,Swirski2012}, least square-based ellipse fitting \cite{Fuhl2015_ExCuSe,Fuhl2016_ElSe}, and circle fitting \cite{Fuhl2016_Microscope}. On contrary to existing handcrafted approaches \cite{Li2005_Starburst,Swirski2012,Fuhl2015_ExCuSe,Fuhl2016_ElSe,Fuhl2016_Microscope}, data-driven approaches were used for pupil localization \cite{PupilNet, PupilNet2} \footnote{We implemented methods in \cite{PupilNet, PupilNet2} to benchmark but we could not reproduce their results. We contacted the authors for their implementation but they mentioned that their codes will not be publicly available until it was published so we only refer to them in the related work section.}. Even though learning visual representations from domain-specific images is a promising alternative to handcrafting representations, training a network from scratch requires more data, time, and hyper-parameter optimization compared to using a pretrained model. Therefore, we proposed a transfer learning-based approach in which we train a single fully connected layer, which only took a few epochs and minutes to converge.

Once pupil locations are obtained, shape-fitting methods can provide pupil sizes as explained in \cite{Gander1994,Lamiroy2007, Chernov2008}. Finally, size variation of right and left pupils can be compared to obtain an RAPD score. To validate and benchmark the performance of RAPD detection algorithms, we need a pupil dataset. Therefore, we analyzed and  reported the main characteristics of existing pupil datasets in Table~\ref{tab:datasets}. The resolution of existing datasets varies between 384x288 and 640x480 (SD). Number of subjects participated in the pupil studies is in between 2 and 40. Most of the datasets \cite{Kasneci2014,Sippel2014,Fuhl2015_ExCuSe,Fuhl2016_ElSe,Tonsen2016_LPW} were obtained with a head-mounted eye tracker whereas a camera system was used in \cite{Swirski2012} and a microscope ocular in \cite{Fuhl2016_Microscope}. Subjects were involved in different activities during the acquisition of these datasets. In Kasneci \cite{Kasneci2014} and Sippel \cite{Sippel2014}, researchers investigated the effect of visual field loss in daily activities and the subjects were either driving or shopping, which also contributed to the majority of images in ExCuSe \cite{Fuhl2015_ExCuSe} and ElSe \cite{Fuhl2016_ElSe} datasets. Activities of the subjects were not specified in Swirski \cite{Swirski2012} and Microscope \cite{Fuhl2016_Microscope} datasets whereas subjects were staring at a moving target in the LPW \cite{Tonsen2016_LPW} dataset. All of the analyzed datasets \cite{Swirski2012,Kasneci2014,Sippel2014,Fuhl2015_ExCuSe,Fuhl2016_ElSe,Fuhl2016_Microscope,Tonsen2016_LPW} provide pupil center position as metadata and pupil ellipses are provided in Swirski \cite{Swirski2012} and LPW \cite{Tonsen2016_LPW}.

The shortcomings of existing pupil datasets are three folds in terms of detecting RAPD. First, the majority of existing pupil datasets are based on static images instead of videos, which makes it difficult or impossible to assess pupillary reflex. Even though Kasneci \cite{Kasneci2014} and Sippel \cite{Sippel2014}  datasets were originally obtained as videos, static images were provided and utilized in pupil detection experiments \cite{Fuhl2015_ExCuSe,Fuhl2016_ElSe}. Labeled pupils in the wild (LPW) is the only analyzed dataset \cite{Tonsen2016_LPW} that provides video sequences. Second, limited control over the acquisition conditions in existing datasets make it unfeasible to assess light reflex. Subjects were exposed to varying stimuli in the acquisition process and environmental conditions are not reported in the datasets. Therefore, it is not possible to isolate and assess the pupillary reflex with respect to the light sources. Third, subjects were not examined for RAPD and there is no corresponding metadata. Therefore, it is not possible to validate RAPD algorithms in these datasets.

\section{RAPD-HD Dataset}
\label{sec:rapd_dataset}
To compensate the shortcomings of existing datasets in terms of limited video data, uncontrolled environments, and lack of pupillary defect annotations, we generated an RAPD dataset with HD video sequences denoted as  \texttt{RAPD-HD} dataset whose main characteristics are summarized in Table~\ref{tab:datasets}. In total, there are 64 test cases with 2 sequences per test case as left and right. Each test case includes multiple light reflex instances. Videos were originally acquired in high definition (HD) resolution and we provide RAPD labels for each video pair as RAPD positive or no RAPD.   
   
\begin{figure}[htbp!]
\vspace{0.1cm}
\begin{minipage}[b]{0.32\linewidth}
  \centering
\includegraphics[width=\textwidth]{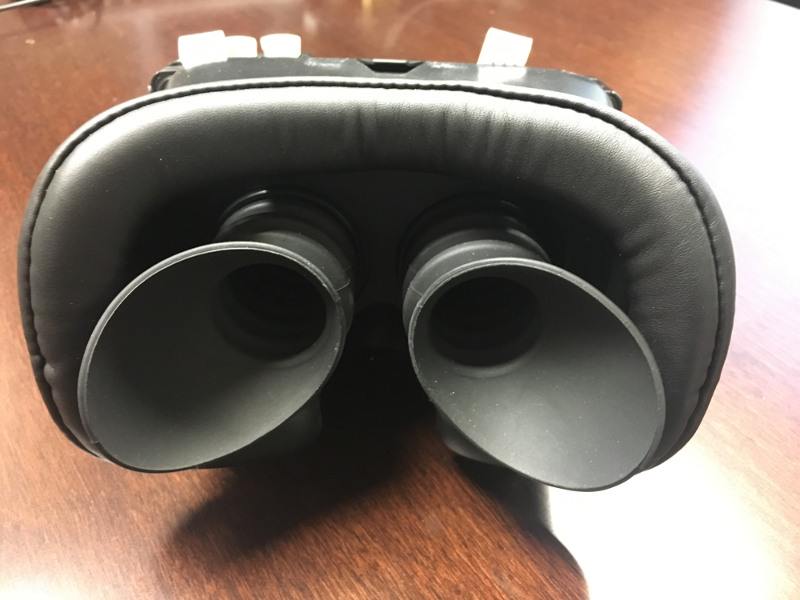}
  \vspace{0.01cm}
  \centerline{\footnotesize{(a) Front view}}%\medskip
  \vspace{0.03 cm}
\end{minipage}
\vspace{0.1cm}
\begin{minipage}[b]{0.32\linewidth}
  \centering
\includegraphics[width=\textwidth]{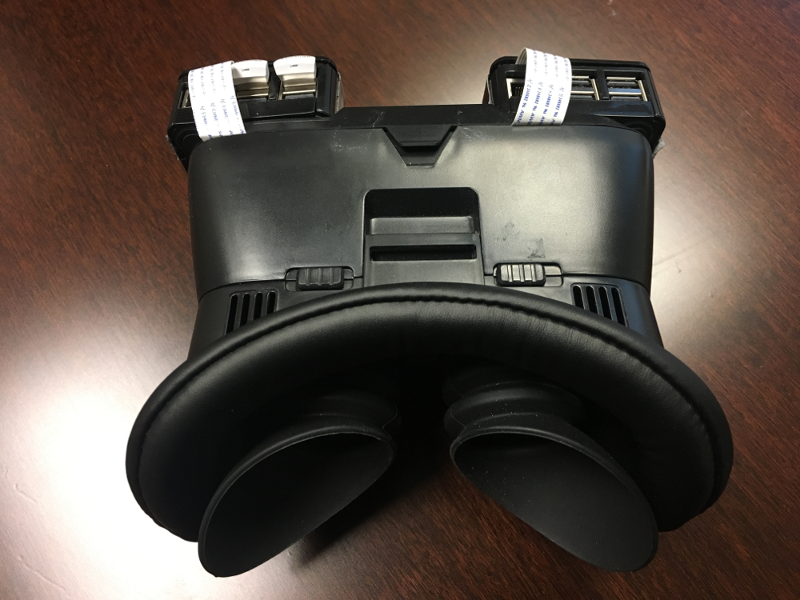}
  \vspace{0.01cm}
  \centerline{\footnotesize{(b) Top view}}%\medskip
  \vspace{0.03 cm}
\end{minipage}
\vspace{0.1cm}
\begin{minipage}[b]{0.32\linewidth}
  \centering
\includegraphics[width=\textwidth]{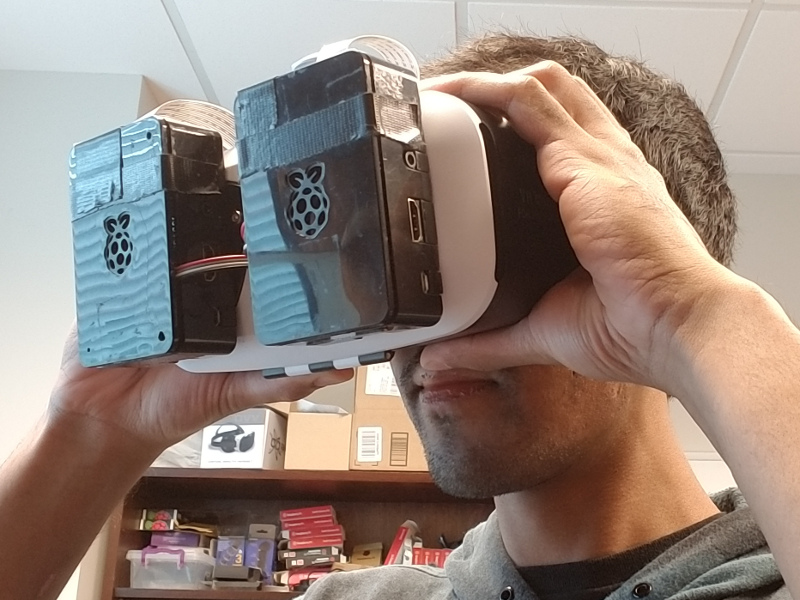}
  \vspace{0.01cm}
  \centerline{\footnotesize{(c) In use}}%\medskip
  \vspace{0.03 cm}
\end{minipage}
\centering
\vspace{-4 mm}
\caption{Developed acquisition device \texttt{lab-on-a-headset}.}
\label{fig:device}
\end{figure}

\vspace{2mm}
\noindent{\bf Acquisition Setup:}~We developed a platform denoted as \texttt{lab-on-a-headset} \cite{Temel2018_RAPD_prov,Temel2019_RAPD_nonprov} that can perform an automated swinging flashlight test as shown in Fig.~\ref{fig:rapd_high_level}. Developed headset, shown in \ref{fig:device}, provides a controlled environment in which each eye has an individual enclosure isolated from external environment. We stimulate eyes independently with a predefined light sequence and record their reactions.

\noindent{\bf Clinical Study:}~We obtained the approvals from the IRB committees of Emory University and Georgia Institute of Technology and started a clinical study at the Grady Memorial Hospital (IRB00099796). RAPD conditions of subjects were diagnosed by the practitioners involved in the clinical study with a swinging flashlight test and a neutral density filter test. The demographics of clinical subjects are summarized in Table \ref{tab:dataset_rapd}. We can observe that average age of RAPD positive subjects are higher than no RAPD subjects. Moreover, all RAPD positive subjects have prior ocular history whereas majority of no RAPD subjects do not have ocular history.

\begin{table}[htbp!]
\vspace{-5mm}
\centering
\caption{Dataset statistics}
\label{tab:dataset_rapd}
\begin{tabular}{ccc}
\hline
\textbf{} & \textbf{No RAPD} & \textbf{RAPD +ve} \\ \hline

\textbf{\begin{tabular}[c]{@{}c@{}}Test cases (\% of total) \end{tabular}}
& \begin{tabular}[c]{@{}c@{}} 32 (50.0\%) \end{tabular}
& \begin{tabular}[c]{@{}c@{}} 32 (50.0\%) \end{tabular}
 \\ \hline

\textbf{\begin{tabular}[c]{@{}c@{}}Male-Female (\% of total) \end{tabular}}
& \begin{tabular}[c]{@{}c@{}} 57.0\%-43.0\% \end{tabular}
& \begin{tabular}[c]{@{}c@{}} 40.0\%-60.0\% \end{tabular}
 \\ \hline

\textbf{\begin{tabular}[c]{@{}c@{}}Age (mean $\pm$ $\sigma$, in years) \end{tabular}}
& \begin{tabular}[c]{@{}c@{}} 49.5 $\pm$ 15.98 \end{tabular}
& \begin{tabular}[c]{@{}c@{}} 58.3 $\pm$ 9.14 \end{tabular}
 \\ \hline

\textbf{\begin{tabular}[c]{@{}c@{}}Prior History (\% of group) \end{tabular}}
& \begin{tabular}[c]{@{}c@{}} 35.7\% \end{tabular}
& \begin{tabular}[c]{@{}c@{}} 100.00\% \end{tabular}
 \\ \hline

\end{tabular}
\vspace{-5 mm}
\end{table}

\section{RAPD Detection Framework}
\label{sec:rapd_framework}
First, we detect the pupils in captured video sequences. Second, we measure relative pupil size variation during the test. Third, we measure the dissimilarity between left and right pupillary reactions to obtain an index between 0 and 1, which is denoted as the \texttt{RAPD index}. Healthy subjects should lead to an \texttt{RAPD index} close to 0 whereas subjects with RAPD positive should lead to higher values. In the following sections, we describe the algorithms that are utilized to perform pupil localization, pupil size measurement, and RAPD assessment.

\begin{table*}[!htbp]
\vspace{-5 mm}
\centering
\caption{Pupil localization methods}
\label{tab:algorithms}
\begin{tabular}{ccccc}
\hline
\textbf{} & \textbf{Starburst}~\cite{Li2005_Starburst} & \textbf{ExCuSe}~\cite{Fuhl2015_ExCuSe} & \textbf{ElSe}~\cite{Fuhl2016_ElSe} &\bf RAPDNet \\ \hline

\textbf{\begin{tabular}[c]{@{}c@{}}Stage 1 \end{tabular}}
& \begin{tabular}[c]{@{}c@{}} Initialize pupil center\\ as image center \end{tabular}
&\begin{tabular}[c]{@{}c@{}} Filtering contours from canny\\ edge maps based on\\ curvature and connected components \end{tabular}
& \begin{tabular}[c]{@{}c@{}} Selection of edges that\\ most likely belong to\\ elliptical edges \\\end{tabular} & \begin{tabular}[c]{@{}c@{}} Classify overlapping patches\\ as pupil and no pupil \end{tabular}
 \\ \hline

\textbf{\begin{tabular}[c]{@{}c@{}}Stage 2 \end{tabular}} & \begin{tabular}[c]{@{}c@{}}Project rays outwardly from\\ pupil center in all directions\\ and mark pixels at high\\ gradients as feature points \end{tabular}  & \begin{tabular}[c]{@{}c@{}}Curve selection using Starburst\\ algorithm to determine feature\\ points of elliptical curves   \end{tabular}   & \begin{tabular}[c]{@{}c@{}} Ellipse fitting and selection\\ on best edges   \end{tabular} & \begin{tabular}[c]{@{}c@{}} Sort the patches based on\\ classification confidence and \\determine top-5 patches \end{tabular}   \\ \hline

\textbf{\begin{tabular}[c]{@{}c@{}}Stage 3 \end{tabular}} &\begin{tabular}[c]{@{}c@{}} Reiteration after updating pupil\\ center based on feature\\ points until convergence \end{tabular} &\begin{tabular}[c]{@{}c@{}} Ellipse fitting on selected curve \end{tabular} & \begin{tabular}[c]{@{}c@{}}  Ellipse evaluation using manually \\ determined ellipse properties that\\ most resemble the pupil \end{tabular} & \begin{tabular}[c]{@{}c@{}}  Obtain the medium location of \\ top-5 patch centers and\\ crop a 60x60 region \end{tabular}     \\ \hline

\textbf{Stage 4} & \begin{tabular}[c]{@{}c@{}}Calculate the mean of\\ all feature points\end{tabular} &\begin{tabular}[c]{@{}c@{}}Obtain ellipse center\\as the pupil center\end{tabular}  &\begin{tabular}[c]{@{}c@{}}Obtain ellipse center\\ as the pupil center\end{tabular} & \begin{tabular}[c]{@{}c@{}} Obtain patch center\\ as the pupil center\end{tabular}   \\ \hline

\end{tabular}
\end{table*}

 \begin{figure*}[htbp!]
\centering
\begin{minipage}[b]{0.85\linewidth}
  \centering
\includegraphics[width=\textwidth]{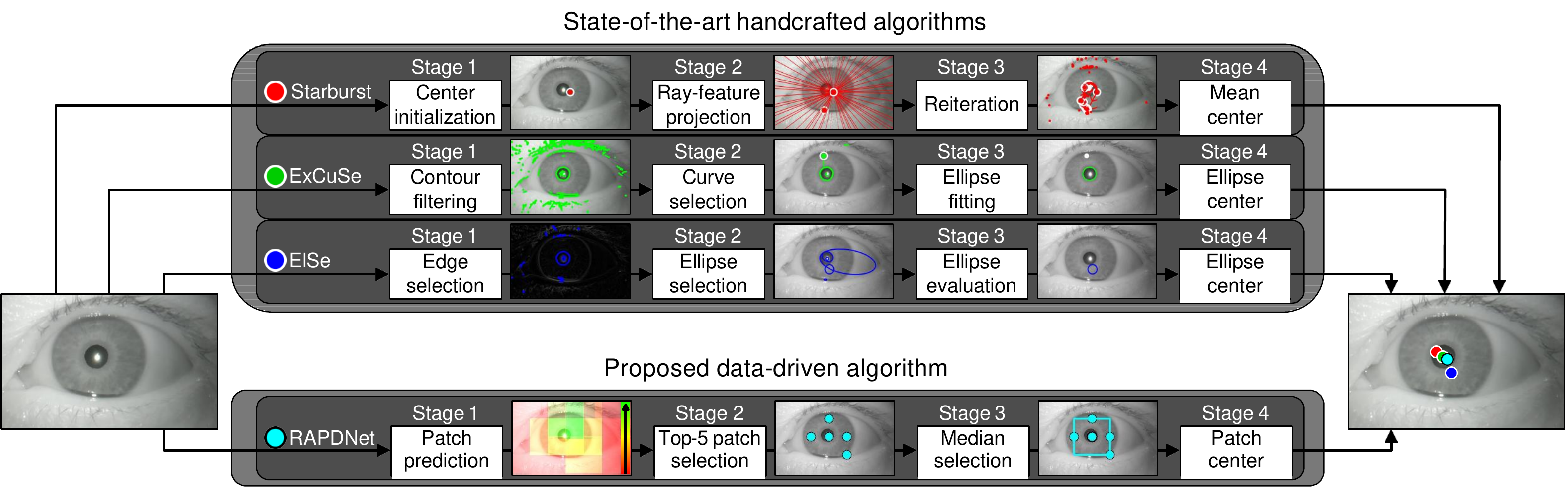}
\end{minipage}
\vspace{0.1cm}
\centering
\vspace{-4 mm}
\caption{Pupil detection pipeline for state-of-the-art handcrafted and proposed data-driven algorithms}
\label{fig:alg_technical}
\vspace{-5 mm}
\end{figure*}

\subsection{Pupil Localization with Handcrafted Algorithms}
\label{subsec:pupil_localization}
We localize pupils with three handcrafted pupil localization algorithms in the literature including Starburst \cite{Li2005_Starburst}, ExCuSe \cite{Fuhl2015_ExCuSe}, and ElSe \cite{Fuhl2016_ElSe}. Main characteristics of these algorithms are summarized in Table \ref{tab:algorithms} along with an illustration of each stage in Fig.~\ref{fig:alg_technical}. 

\noindent{\bf Starburst \cite{Li2005_Starburst}:}~Pupil center is initialized as the image center in Stage 1 and rays are projected outwardly from the center to detect edge features via gradients in Stage 2. Then, pupil center is updated with mean feature location as shown with white circle from which rays are projected. These pupil center update and ray projection procedures are iterated until all the gradients are swept in Stage 3. Finally, pupil center is estimated based on mean location of the eventual feature points.

\noindent{\bf ExCuSe \cite{Fuhl2015_ExCuSe}:}~Curved segments are obtained from Canny edge maps which are filtered to exclude thin lines and small rectangular surfaces in Stage 1. Then, longest curve that encloses the darkest area is selected to project a ray between the curve and the image center in Stage 2. After the ray projection, an ellipse is fit around the estimated pupil in Stage 3 whose center corresponds to the pupil location.

\noindent{\bf ElSe \cite{Fuhl2016_ElSe}:}~Canny edge maps are extracted based on the edges split around orthogonal connections with more than two neighbors. These detected edges are thinned and straightened to enhance curved segment selection. Then, ellipse fitting is performed over detected edges and a subset of ellipses are selected based on the intensity within elliptic regions and their width-height ratio. Finally, the ellipse with the lowest inner gray values and a width-height ratio close to one is selected to obtain the pupil center estimate.

\subsection{Pupil Localization based on Transfer Learning}
\label{sec:framework_data_driven}
Visual representations learned by state-of-the-art object recognition models include low level characteristics based on edges and shapes. We cannot directly use these models for localizing or detecting pupils in eye images because they are originally trained for generic object recognition tasks in large-scale datasets such as ImageNet \cite{ImageNet}. However, the visual representations learned by these models can still be useful for recognizing and localizing eye structures. In this study, we focus on convolutional neural networks (CNNs), which include convolutional layers that learn visual representations and fully connected layers that map these visual representations to target classes. Specifically, we use the AlexNet \cite{AlexNet} architecture that is based on five convolutional layers and three fully connected layers. Size of the convolutional layers vary between 11x11, 5x5, and 3x3. 

\begin{figure*}[tbp!]
\centering
\begin{minipage}[b]{0.85\linewidth}
  \centering
\includegraphics[width=\textwidth]{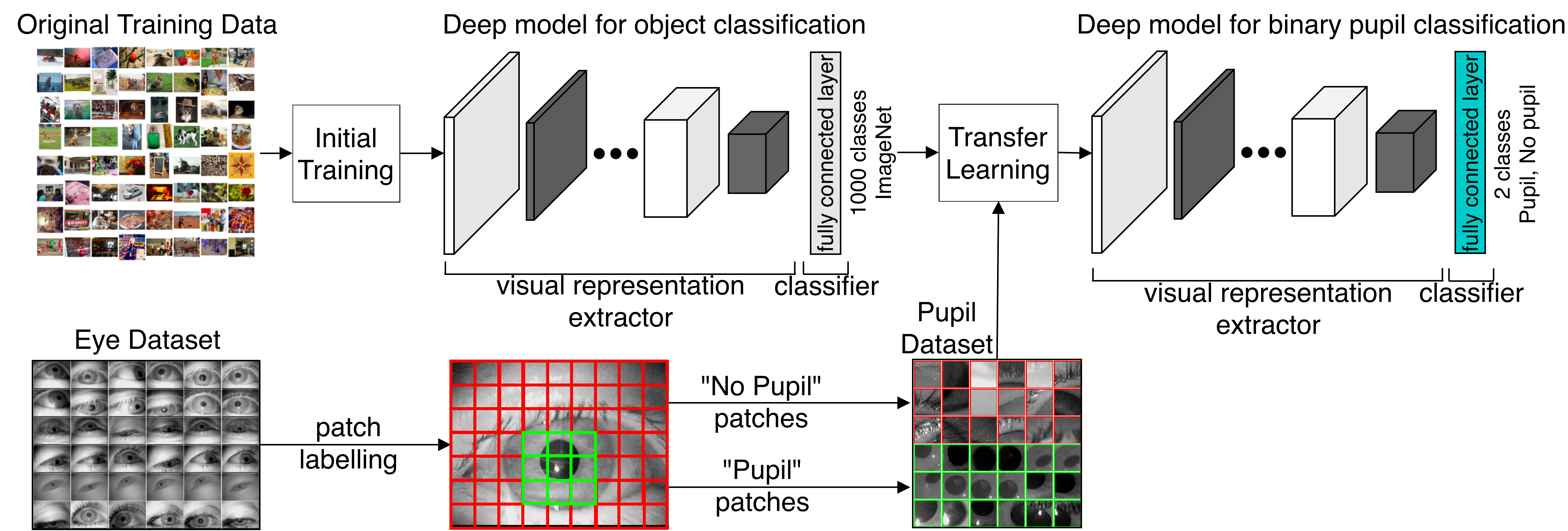}
\end{minipage}
\vspace{0.1cm}
\centering
\vspace{-2 mm}
\caption{Transfer learning framework for obtaining a pupil detection algorithm based on generic object recognition architectures.}
\label{fig:transfer}
\vspace{-2 mm}
\end{figure*}

In this study, we use a pretrained AlexNet \cite{torchvision2019}, which was trained with ImageNet to classify generic objects into 1,000 classes. We transformed this object recognition model into a pupil detector with transfer learning as shown in Fig.~\ref{fig:transfer}. Specifically, we keep the convolutional layers that output visual representations and exclude all  the pretrained fully connected layers at the end of AlexNet. Then, we added a single fully connected layer to map visual representations in the final layer to two classes as pupil and no pupil. In the training, we froze all the convolutional layers and trained the fully connected layer with image patches labeled as pupil and no pupil. We used a total of 3,620 images from datasets XII, X, XIII, XIV, XVI, XVII, and XVIII in the ElSe study \cite{Fuhl2016_ElSe} for training, validation, and testing. We visually inspected all 24 datasets in the ElSe study and selected the ones without lens and glare artifact. Data was split into 60\% training set, 20\% validation set, and 20\% test set. In the ElSe dataset, a patch of 50x50 is wide enough to cover pupil region so we split the images into 50x50 overlapping patches with a stride of half the patch size to increase the number of patches. We labeled the patches that include pupil center as pupil images and the remaining ones as no pupil images. To balance the training and test set, we used a subset of no pupil images whose number is equal to pupil images. Moreover, we experimented with different patch sizes including 75, 100, 125, and 150.

Transfer learning of AlexNet resulted in a pupil detection model denoted as \texttt{RAPDNet}. To localize pupils in an image, we scan the image with overlapping patches and classify each patch as pupil or no pupil with \texttt{RAPDNet}. After obtaining a class for each patch, we sort the pupil patches based on their classification confidence and compute the median location of the top-5 pupil patches with the highest confidence. In recognition tasks, top-5 accuracy is a commonly used metric, which considers recognition as correct if the target class is among the top-5 estimates. Similarly, we look at the top-5 estimates but instead of focusing on one estimate, we obtain the median location of all 5 estimates. We proposed a patch-based localization instead of bounding box-based detection because dataset annotations only included pupil centers. Main characteristics of the proposed algorithm are summarized in Table \ref{tab:algorithms} along with an illustration of each stage in Fig.~\ref{fig:alg_technical}.

\subsection{Pupil Size Measurement}
\label{subsec:pupil_size_measurement}
We perform pupil size measurement over image patches automatically cropped around the estimated pupil center. In the first crop configuration \emph{image size/2}, estimated pupil region is cropped to half of the downsampled input image size. In the second crop configuration \emph{60x60}, a 60 by 60 patch is cropped around estimated pupil center. We detect and measure pupils with Circular Hough Transform,  which transforms points in the 2-dimensional image plane into circular cones in the Hough space. Image points in the same circle will have cones in the Hough space that intersect at points, and these intersection points will indicate pupil size measurements.

In this study, we use the Hough gradient method to perform circular Hough transform (CHT) \cite{opencv_library}. The number of detected circles per image is limited to one because there is at most one pupil in the field of view. Search range for radius is set to 5-30 pixels, which cover the average pupil range between 2mm and 8mm \cite{Spector1990}. In CHT, there are two threshold parameters denoted as Canny threshold and accumulator threshold. Canny is the edge detection threshold and accumulator threshold corresponds to the detection confidence. Threshold values can be manually fine tuned according to target applications. However, in this study, we automate the threshold selection process by sweeping a range of values. For Canny threshold, we set the maximum sweep value to 255 (maximum gradient value of an 8-bit image) and the minimum sweep value to mean of the image. For accumulator threshold, the maximum sweep value is set to 100, which corresponds to the maximum accumulator value. Automated threshold selection starts from the maximum sweep values and iterates to lower values until a circle is detected.

\subsection{RAPD Assessment}
We measure pupil size variations in swinging flashlight video sequences by combining pupil localization algorithms with pupil size measurement methods, which leads to baseline algorithms. Pupillary light reflex measurements are processed with a median filter to eliminate outliers. Moreover, we also experiment with an additional moving average step in Section~\ref{sec:experiments} to analyze the contribution of smoothing out signals. To measure dissimilarity between pupillary reflexes, we propose a dissimilarity index denoted as $RAPD~index$, which is formulated as   
\begin{equation}
    1.0-\frac{min(|\Delta_R|,|\Delta_L|)}{max(|\Delta_R|,|\Delta_L|)},    
\end{equation}
where $\Delta_R$ is the pupil size change in the right eye, $\Delta_L$ is the pupil size change in the left eye, $|~|$ is the absolute value operator, $min$ is the minimum operator, and $max$ in the maximum operator. $RAPD~index$ is 0.0 when compared pupil size changes are identical whereas it converges to 1.0 as pupil size changes differ from each other. In addition to $RAPD~index$, we also use 1.0 minus correlation metrics in the Section~\ref{sec:experiments} to compare the performance of alternative metrics. We need a numerical threshold to determine the RAPD status of the subjects based on the $RAPD~index$. To determine the classification threshold for all the benchmarked algorithms, we measure the sensitivity and specificity for all possible threshold values and select the values that correspond to highest sensitivity and specificity for each algorithm.

\section{Experiments and Discussion}
\label{sec:experiments}
At first, we analyze the classification performance of the proposed pupil detection algorithm in Section~\ref{subsec:experiments_pupil}. Then, we report and analyze the performance of RAPD detection algorithms including benchmarked hancrafted algorithms and proposed data-driven algorithm \texttt{RAPDNet} in  Section~\ref{subsec:experiments_RAPD}.

\begin{figure}[htbp!]
%\vspace{-3 mm}
 \centering
\begin{minipage}[b]{0.4\linewidth}
  \centering
\includegraphics[width=\textwidth]{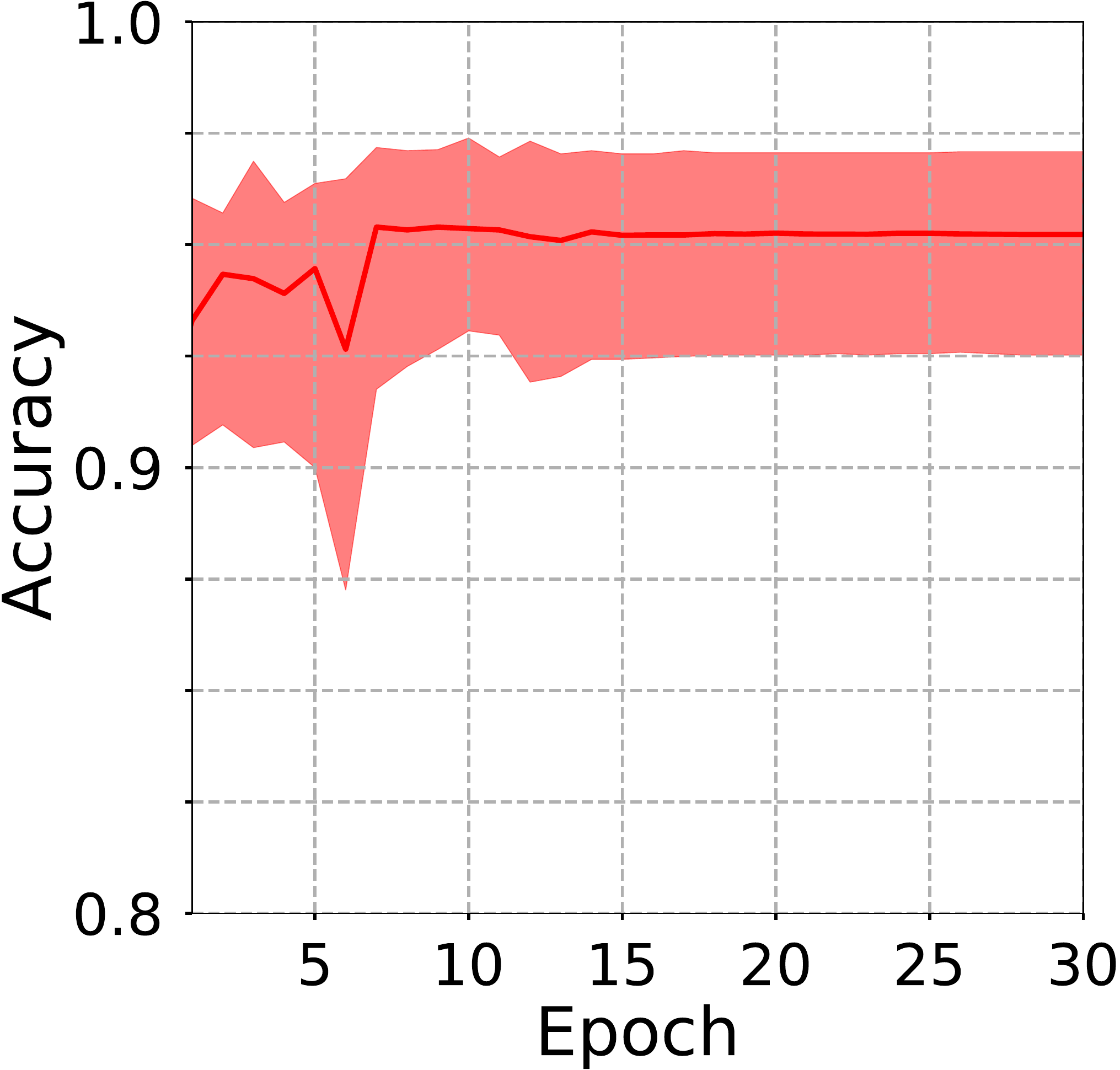}
  \vspace{0.01cm}
  \centerline{\footnotesize{(a) Validation accuracy}}%\medskip
  \vspace{0.03 cm}
\end{minipage}
\vspace{0.1cm}
\begin{minipage}[b]{0.4\linewidth}
  \centering
\includegraphics[width=\textwidth]{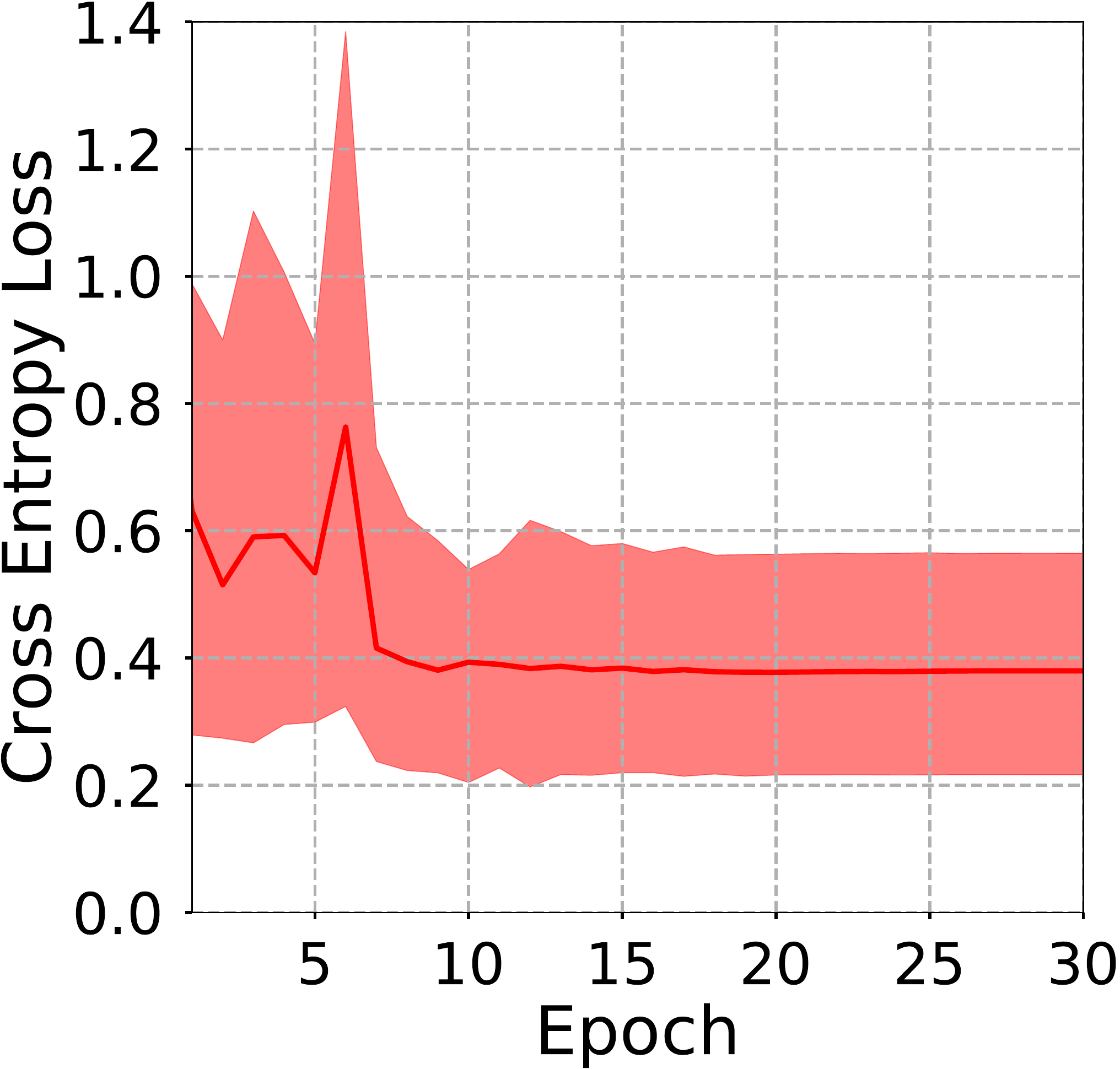}
  \vspace{0.01cm}
  \centerline{\footnotesize{(b) Validation loss}}%\medskip
  \vspace{0.03 cm}
\end{minipage}
\centering
\vspace{-3 mm}
\caption{Pupil recognition accuracy and loss of the developed algorithm based on transfer learning with respect to training epochs.}
\label{fig:recognitionresults}
\vspace{-4 mm}
\end{figure}

\subsection{Pupil Recognition}
\label{subsec:experiments_pupil}
We report the classification performance of the developed pupil recognition algorithm in terms of accuracy and cross entropy loss in Fig.~\ref{fig:recognitionresults}. In these plots, x-axis corresponds to the epoch index and y-axis corresponds to the performance metrics. In the training experiments, we tried different patch sizes including 50, 75, 100, 125, and 150. We combined the results of the models trained with different patch sizes. For example, at a specific epoch, minimum and maximum values in the plot are obtained from the minimum and the maximum performance of these five patch configurations. We can observe that classification performance variations attenuate after first eight epochs and converge to final validation performance. 

With 60 million parameters and 650,000 neurons, training AlexNet would take many epochs until convergence. However, in the proposed algorithm, we use the convolutional features as they are, remove the pretrained fully connected layers, add a single fully connected layer with two output classes at the end, and only train this single layer. Given that we only train a single layer and the task is binary classification, the algorithm converges in a few epochs as shown in Fig.~\ref{fig:recognitionresults}. Not only the algorithm converges in a few epochs, but it also leads to a 90+\% accuracy after the first epoch. In order for an algorithm to reach high accuracies after one epoch and to converge after few epochs, the pretrained features should be a very good fit for the target application. Visual representations obtained from AlexNet include curvy and linear patches that can be transposed and weighted to represent a circular region and the intensity information can be used to differentiate the darker pupil region from the surrounding. Because of the direct usability of the pretrained feature representations, we were able to generalize well through transfer learning.

\begin{table}[htbp!]
\vspace{-4mm}
\centering
\caption{{Validation and test accuracy of the developed pupil recognition algorithm based on transfer learning.}}\begin{tabular}{ccccc}
\hline
              \textbf{50x50} & \textbf{75x75} & \textbf{100x100} &\textbf{125x125}&\textbf{150x150} \\ \hline

\multicolumn{5}{c}{\textbf{Validation Set}} \\ \hline
             
\cellcolor{yellow!30}\bf 0.974          & 0.965        & 0.963     & 0.952          & 0.931             \\ \hline

\multicolumn{5}{c}{\textbf{Test Set}} \\ \hline

\cellcolor{yellow!30}\bf 0.957          & 0.954        & 0.954     & 0.948          & 0.933              \\ \hline

\end{tabular}
\label{tab:recognition}
\end{table}

After the training stage, we test the models over the final test set and obtain their performance as reported in Table~\ref{tab:recognition} in which we highlight the highest classification performance with a yellow background and a bold font. Overall, we can achieve an accuracy of $0.974$ in the validation set and an accuracy of $0.957$ in the test set. Based on the results reported in Table~\ref{tab:recognition}, 50x50 patch size leads to the highest classification accuracy in the validation set as well as in the test set. In the \texttt{lab-on-a-headset} setup, 60 pixel corresponds to a physical distance slightly more than 9mm, which is wide enough to cover dilated pupils. A patch size significantly wider than pupil size can lead to more false positives. Therefore, we used a patch of 60x60 in the \texttt{RAPD-HD} dataset experiments, which is re-scaled to 50x50 before feeding into the trained models. We set the stride as half of the patch size to be consistent with the training setup.

\subsection{RAPD Detection}
\label{subsec:experiments_RAPD}
 We describe the detection performance metrics in Table~\ref{tab:metrics} and report the results in Table \ref{tab:rapd_benchmark}.
 We obtain the area under curve (AUC) values for receiver-operating-characteristic (ROC) plots by sweeping the detection threshold and measuring the swept area. For each localization algorithm, we report the results for two patch sizes and smoothing configurations. Detection threshold was set separately for each algorithm to maximize sensitivity and specificity. We highlight the algorithm results with bold font and yellow background that correspond to the highest detection performance for each performance metric in Table \ref{tab:rapd_benchmark}. Out of 9 performance metrics, \texttt{RAPDNet} leads to the highest performance in all the performance categories. In the remaining of this study, we utilize the patch size and smoothing configuration that leads to the highest AUC in ROC analysis for each algorithm. Specifically, we use moving average for all the algorithms, and we utilize half image size as the patch size for all the algorithms other than Starburst.
\begin{table}[htbp!]
\vspace{-4 mm}
\centering
\caption{Detection performance metrics.}
\begin{tabular}{c|c}
\hline
\textbf{Term} & \textbf{Description} \\ \hline
Positive ($P$) & Number of RAPD positive cases \\ 
Negative ($N$) & Number of cases without RAPD \\ 
True positive ($TP$) & Number of correct RAPD detections \\ 
True negative ($TN$) & Number of  correct no RAPD detections \\ 
False positive ($FP$) & Number of false RAPD detection \\ 
False negative($FN$) & Number of undetected RAPD cases\\ 
 Area under Curve ($AUC$)& AUC for ROC plot  \\ \hline
\textbf{Term} & \textbf{Formulation} \\ \hline
Precision & $TP/(TP+FP)$ \\
Sensitivity/Recall/TPR & $TP/(TP+FN)$ \\
FPR & $FP/(TN+FP)$ \\
Specificity/TNR & $TN/(TN+FP)$ \\
FNR & $FN/(TP+FN)$ \\
Accuracy & $(TP+TN)/(TP+TN+FN+FP)$ \\
Balanced Accuracy &$(sensitivity +specificity)/2$  \\
% Balanced Accuracy &(TP/P+TN/N)/2  \\
\multirow{2}{*}{\begin{tabular}[c]{@{}c@{}}$f_1$ score\end{tabular}} &\multirow{2}{*}{\begin{tabular}[c]{@{}c@{}}$\frac{2 \cdot precision \cdot recall}{ precision+recall}$\end{tabular}}   \\ \\ \hline
 \end{tabular}
\label{tab:metrics}
\end{table}

\begin{table*}[htbp!]
\centering
\footnotesize
\caption{RAPD detection benchmark.}
\label{tab:rapd_benchmark}
\begin{tabular}{ccc|ccccccccc}
\hline
 \multirow{2}{*}{\textbf{\begin{tabular}[c]{@{}c@{}}Localization\\Algorithm\end{tabular}}}&\multirow{2}{*}{\textbf{\begin{tabular}[c]{@{}c@{}}Patch\\Size\end{tabular}}} &\multirow{2}{*}{\textbf{\begin{tabular}[c]{@{}c@{}}Smoothing\end{tabular}}} &\multirow{2}{*}{\textbf{\begin{tabular}[c]{@{}c@{}}Precision\end{tabular}}}&\multirow{2}{*}{\textbf{\begin{tabular}[c]{@{}c@{}}Sensitivity \\ (TPR)\end{tabular}}}&\multirow{2}{*}{\textbf{\begin{tabular}[c]{@{}c@{}}FPR\end{tabular}}}&\multirow{2}{*}{\textbf{\begin{tabular}[c]{@{}c@{}}Specificity\\ (TNR)\end{tabular}}}&\multirow{2}{*}{\textbf{\begin{tabular}[c]{@{}c@{}}FNR\end{tabular}}}&\multirow{2}{*}{\textbf{\begin{tabular}[c]{@{}c@{}}Accuracy\end{tabular}}} &\multirow{2}{*}{\begin{tabular}[c]{@{}c@{}}$\bm{f_1}$\end{tabular}}& \multirow{2}{*}{\begin{tabular}[c]{@{}c@{}}$\bm{AUC_{PR}}$\end{tabular}}  &\multirow{2}{*}{\begin{tabular}[c]{@{}c@{}}$\bm{AUC_{ROC}}$\end{tabular}}  \\ &&&&&&&&&&&
 \\ \hline
 
 %\multirow{2}{*}{\textbf{\begin{tabular}[c]{@{}c@{}}-\end{tabular}}}  &full&-&58.3&61.8&31.3&68.8&38.2&65.8&65.3&0.600&0.646 \\
 %&full&mov avg&63.6&61.8&25.0&75.0&38.2&69.5&68.4&0.627&0.672 \\    \hline
 
 %\cellcolor{yellow!30}\bf
 
 \multirow{4}{*}{\textbf{\begin{tabular}[c]{@{}c@{}}Starburst\end{tabular}}}  &\multirow{2}{*}{\begin{tabular}[c]{@{}c@{}}image size/2\end{tabular}}&-&78.6 &	68.8&	18.8&	81.3&	31.3&	75.0&	0.733 &	0.844 &	0.789 \\
 &&mov avg&84.4&	84.4&	15.6&	84.4&	15.6&	84.4&	0.844 &	0.885 &	0.844 \\  
 &\multirow{2}{*}{\begin{tabular}[c]{@{}c@{}}60x60\end{tabular}}&-&87.5&	65.6&	9.4&	90.6&	34.4&	78.1&	0.750&	0.860	&0.795 \\
 &&mov avg&88.9&	75.0&	9.4&	90.6&	25.0&	82.8&	0.814&	0.894&	0.848 \\   \hline
 
 \multirow{4}{*}{\textbf{\begin{tabular}[c]{@{}c@{}}ExCuSe\end{tabular}}}  &\multirow{2}{*}{\begin{tabular}[c]{@{}c@{}}image size/2\end{tabular}}&-&75.8&	78.1&	25.0&	75.0&	21.9&	76.6&	0.769&	0.741&	0.749 \\
 &&mov avg&78.8&	81.3&	21.9&	78.1&	18.8&	79.7&	0.800&	0.762&	0.780 \\  
 &\multirow{2}{*}{\begin{tabular}[c]{@{}c@{}}60x60\end{tabular}}&-&75.9&	68.8&	21.9&	78.1&	31.3&	73.4&	0.721&	0.640&	0.691 \\
 &&mov avg&74.3&	81.3&	28.1&	71.9&	18.8&	76.6&	0.776&	0.644&	0.704 \\   \hline
 
  \multirow{4}{*}{\textbf{\begin{tabular}[c]{@{}c@{}}ElSe\end{tabular}}}  &\multirow{2}{*}{\begin{tabular}[c]{@{}c@{}}image size/2\end{tabular}}&-&70.3&	81.3&	34.4&	65.6&	18.8&	73.4&	0.754&	0.654&	0.709 \\
 &&mov avg&76.5&	81.3&	25.0&	75.0&	18.8&	78.1&	0.788&	0.676&	0.744 \\  
 &\multirow{2}{*}{\begin{tabular}[c]{@{}c@{}}60x60\end{tabular}}&-&75.0&	75.0&	25.0&	75.0&	25.0&	75.0&	0.750&	0.655&	0.730 \\
 &&mov avg&78.1&	78.1&	21.9&	78.1&	21.9&	78.1&	0.781&	0.670&	0.743\\   \hline

  \multirow{4}{*}{\textbf{\begin{tabular}[c]{@{}c@{}}RAPDNet\end{tabular}}}  &\multirow{2}{*}{\begin{tabular}[c]{@{}c@{}}image size/2\end{tabular}}&-&90.3&	87.5&	9.4&	90.6&	12.5&	89.1&	0.889&	0.940&	0.914 \\
 &&mov avg&90.6&	90.6&	9.4&	90.6&	9.4&	90.6&	0.906&\cellcolor{yellow!30}\bf	0.956&\cellcolor{yellow!30}\bf	0.929 \\  
 &\multirow{2}{*}{\begin{tabular}[c]{@{}c@{}}60x60\end{tabular}}&-&\cellcolor{yellow!30}\bf93.5&	90.6&	\cellcolor{yellow!30}\bf6.3&\cellcolor{yellow!30}\bf	93.8&	9.4&\cellcolor{yellow!30}\bf	92.2&\cellcolor{yellow!30}\bf	0.921&	0.930&	0.918 \\
 &&mov avg&85.7&\cellcolor{yellow!30}\bf	93.8&	15.6&	84.4&\cellcolor{yellow!30}\bf	6.3&	89.1&	0.896&	0.907&	0.902 \\   \hline
 \end{tabular}
\vspace{-4mm}
\end{table*}

\begin{figure}[htbp!]
\vspace{-7mm}
\centering
\vspace{0.1cm}
\vspace{4 mm}
\begin{minipage}[b]{0.48\linewidth}
  \centering
\includegraphics[width=\textwidth]{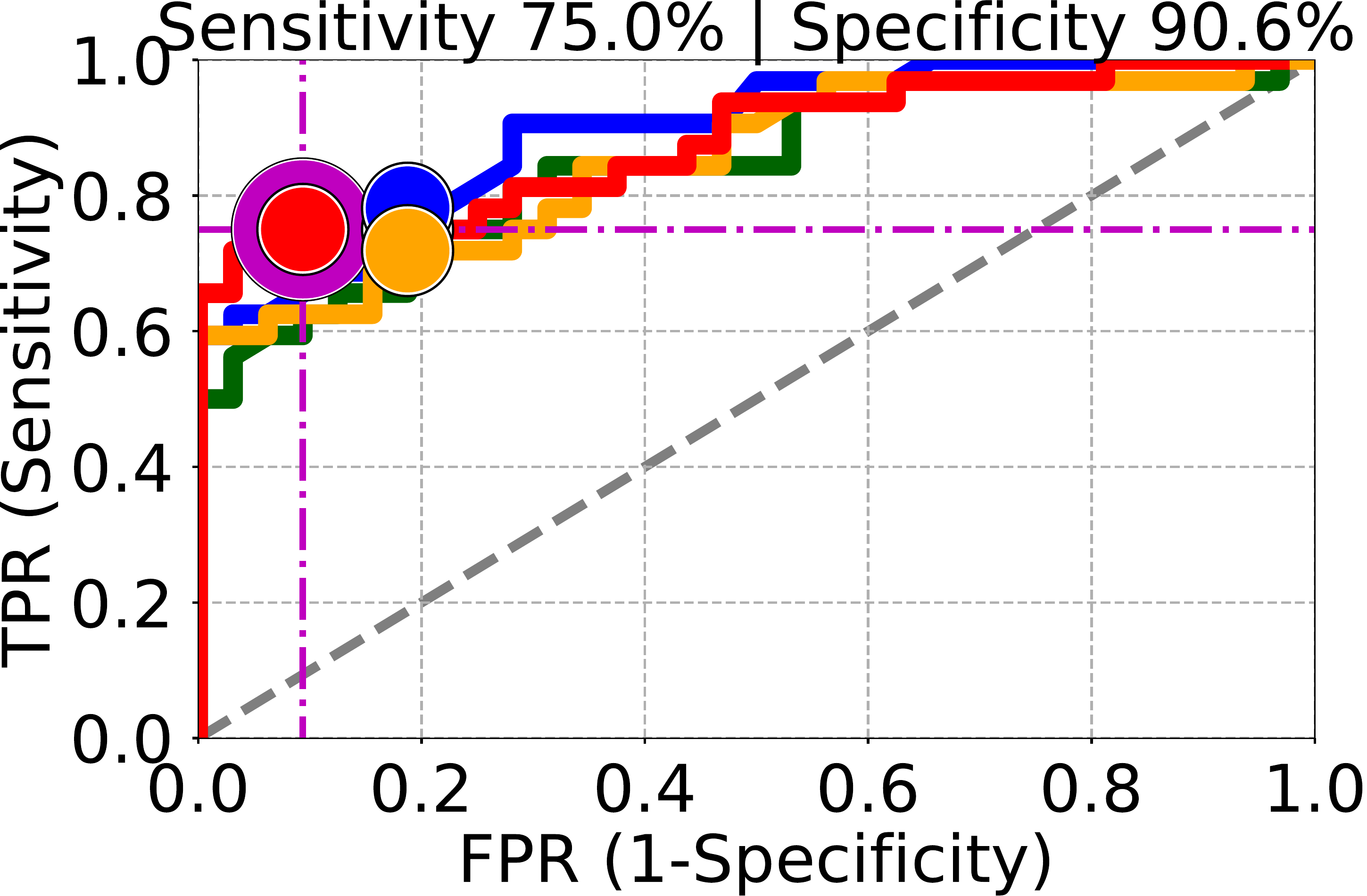}
  \vspace{0.01cm}
  \centerline{\footnotesize{(a) Starburst ROC}}%\medskip
  \vspace{0.03 cm}
\end{minipage}
\vspace{0.1cm}
\begin{minipage}[b]{0.48\linewidth}
  \centering
\includegraphics[width=\linewidth]{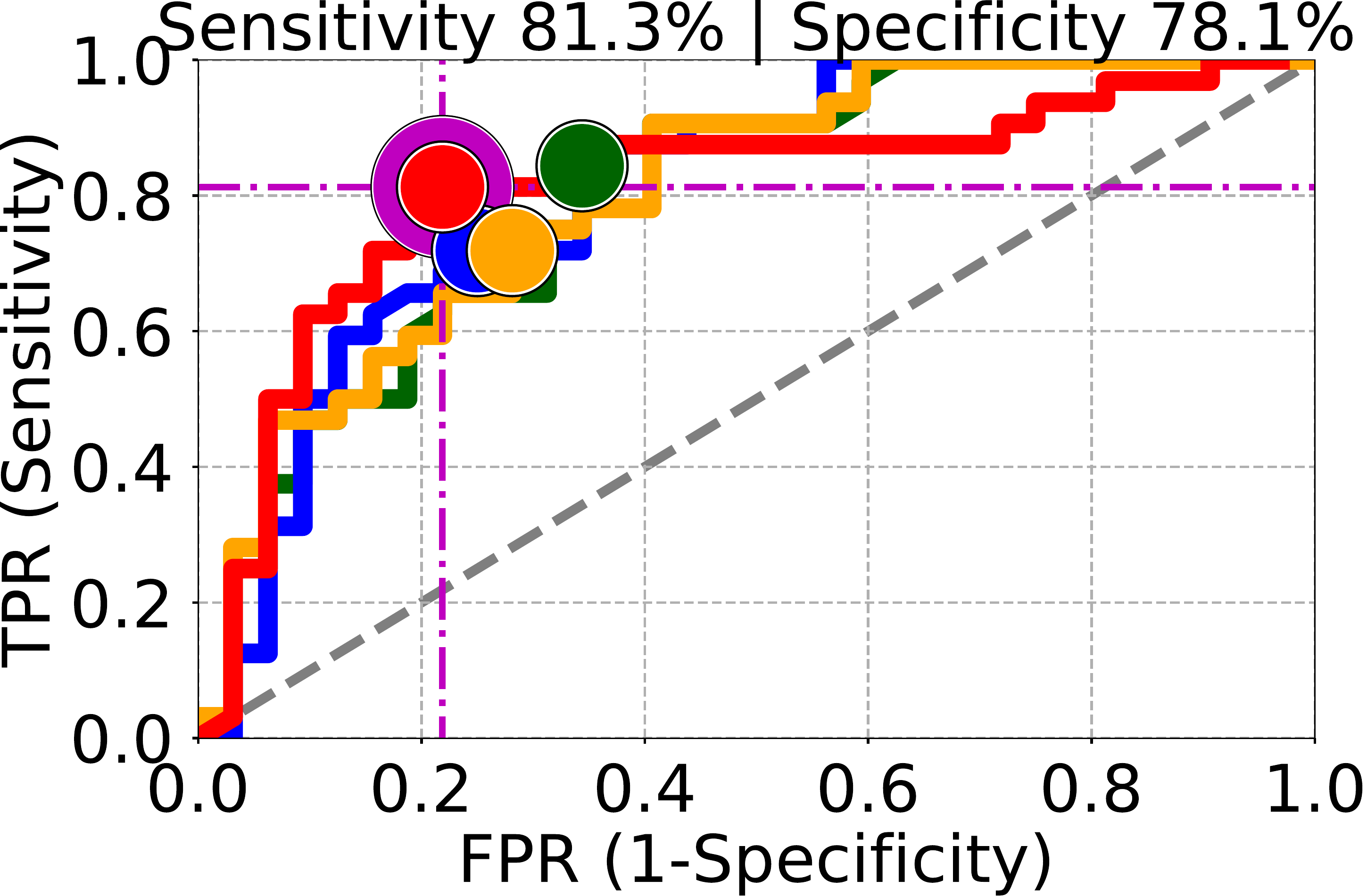}
  \vspace{0.01 cm}
  \centerline{\footnotesize{(b) ExCuSe ROC} }%\medskip
  \vspace{0.03 cm}
\end{minipage}
\vspace{0.1cm}
\begin{minipage}[b]{0.48\linewidth}
  \centering
\includegraphics[width=\textwidth]{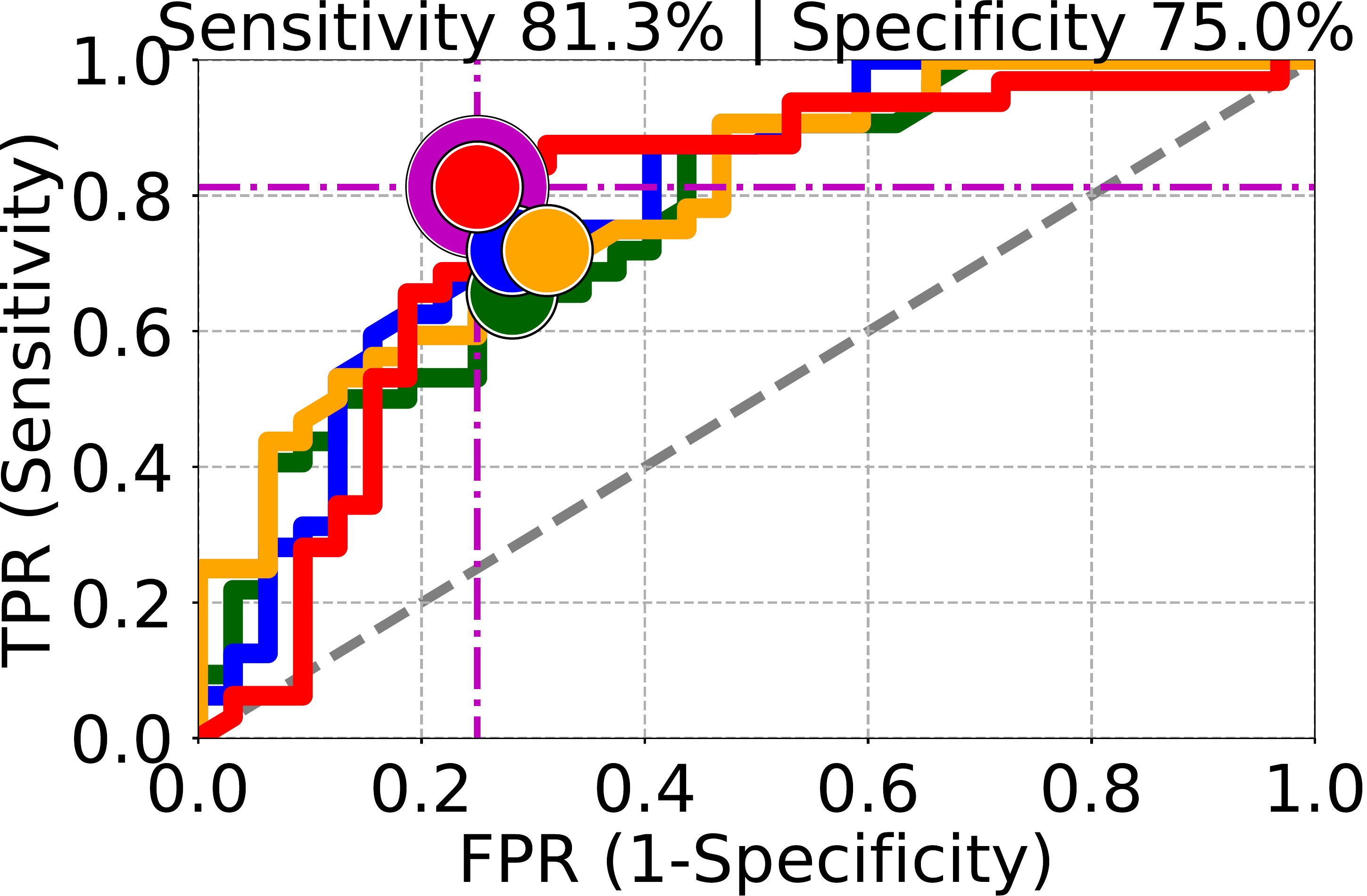}
  \vspace{0.01cm}
  \centerline{\footnotesize{(c) ElSe ROC}}%\medskip
  \vspace{0.03 cm}
\end{minipage}
\vspace{0.1cm}
\begin{minipage}[b]{0.48\linewidth}
  \centering
\includegraphics[width=\textwidth]{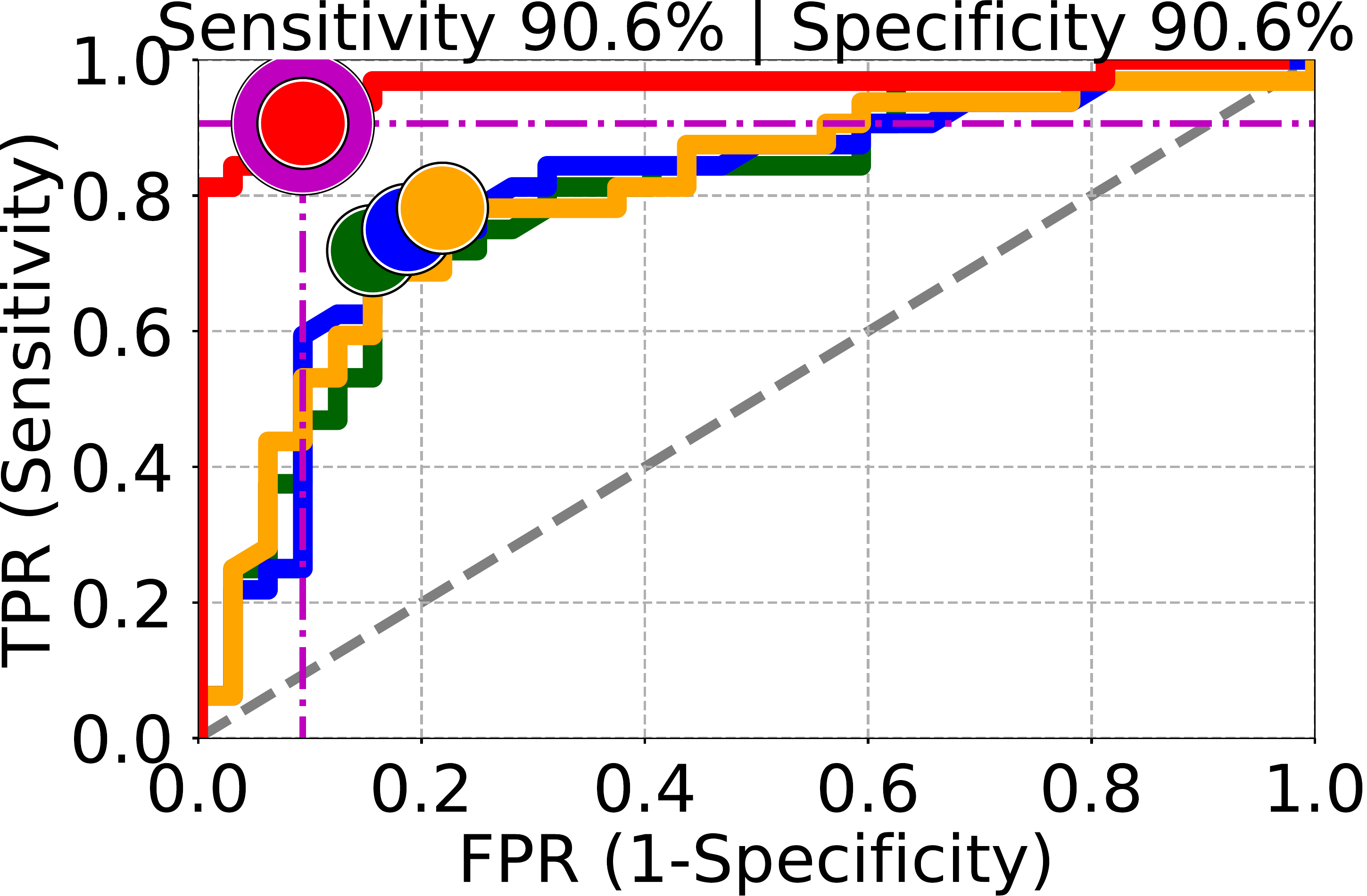}
  \vspace{0.01cm}
  \centerline{\footnotesize{(d) RAPDNet ROC}}%\medskip
  \vspace{0.03 cm}
\end{minipage}

\begin{minipage}[b]{0.98\linewidth}
  \centering
\includegraphics[width=\linewidth]{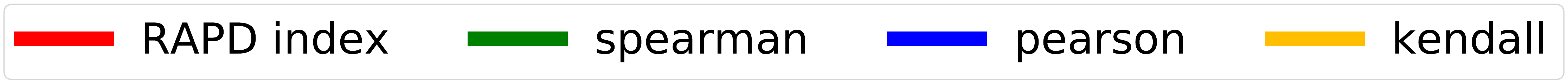}
\end{minipage}

\centering
%\vspace{-5 mm}
\caption{ROC curves for RAPD detection algorithms.}
\label{fig:AUCs}
\vspace{-2 mm}

\end{figure}

\begin{figure}[tbp!]
\centering
\vspace{0.1cm}
\begin{minipage}[b]{0.48\linewidth}
  \centering
\includegraphics[width=\textwidth]{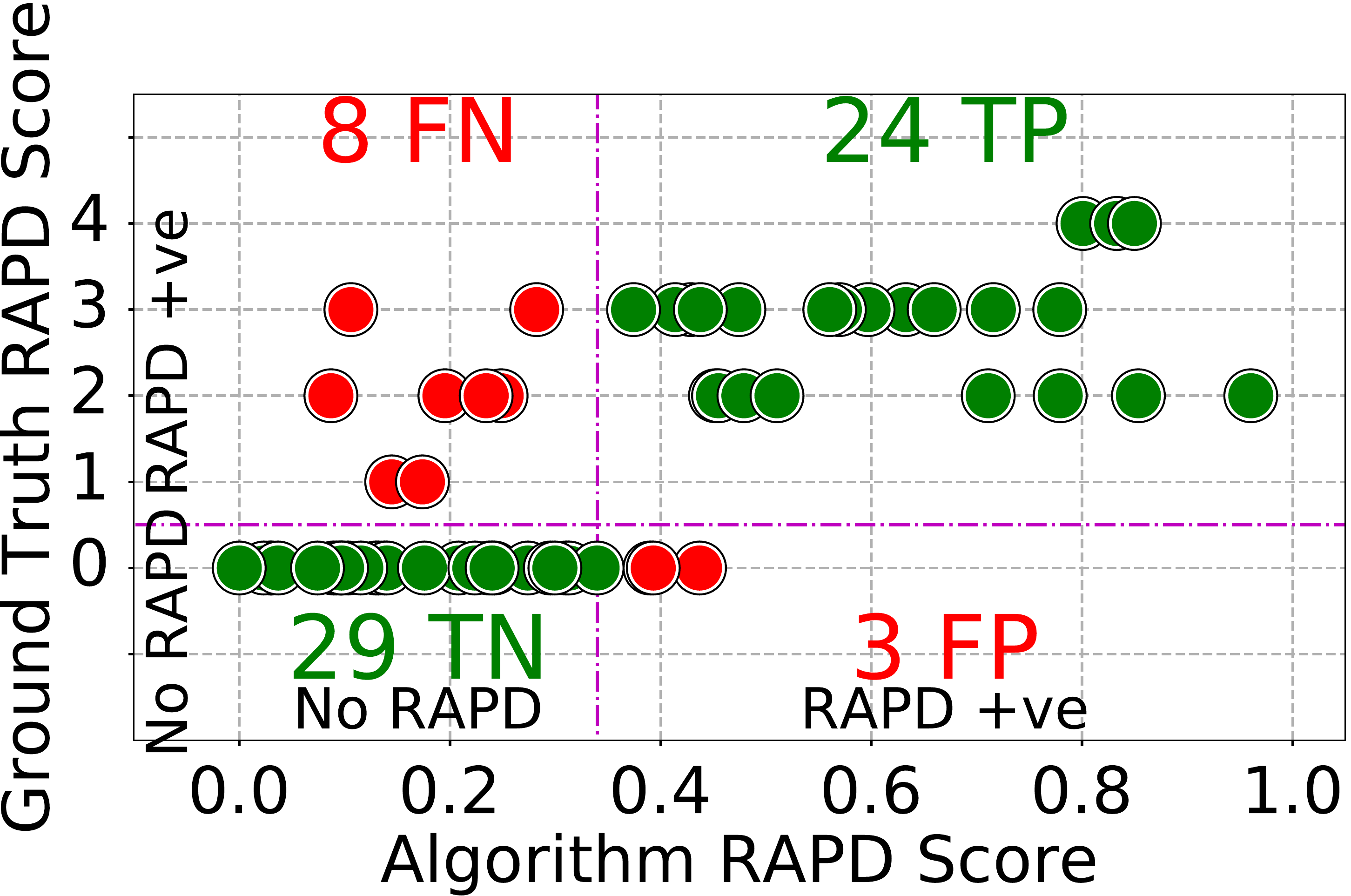}
  \vspace{0.01cm}
  \centerline{\footnotesize{(a) Starburst Scatter}}%\medskip
  \vspace{0.03 cm}
\end{minipage}
\vspace{0.1cm}
\begin{minipage}[b]{0.48\linewidth}
  \centering
\includegraphics[width=\linewidth]{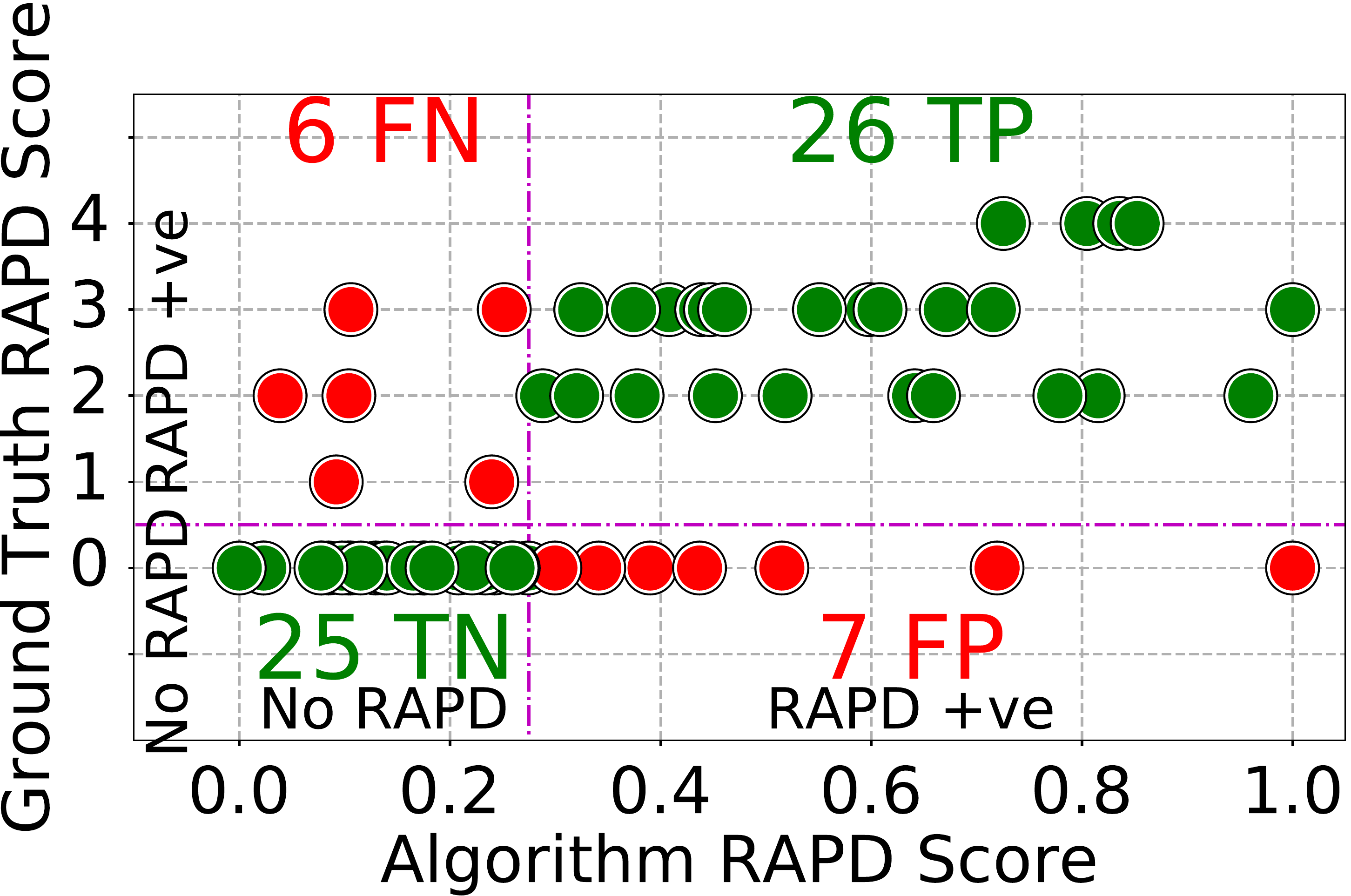}
  \vspace{0.01 cm}
  \centerline{\footnotesize{(b) ExCuSe Scatter} }%\medskip
  \vspace{0.03 cm}
\end{minipage}
\vspace{0.1cm}
\begin{minipage}[b]{0.48\linewidth}
  \centering
\includegraphics[width=\textwidth]{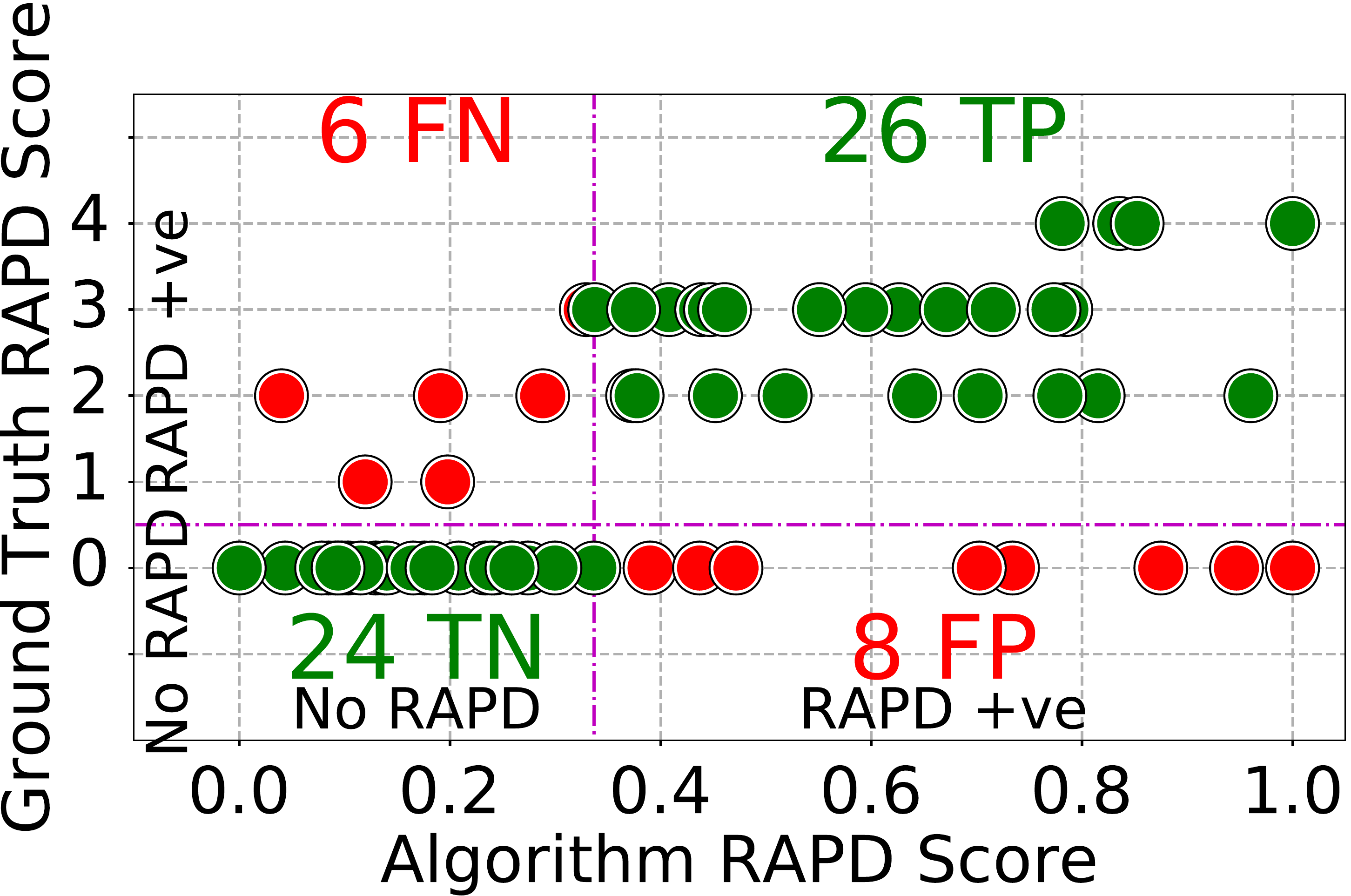}
  \vspace{0.01cm}
  \centerline{\footnotesize{(c) ElSe Scatter}}%\medskip
  \vspace{0.03 cm}
\end{minipage}
\vspace{0.1cm}
\begin{minipage}[b]{0.48\linewidth}
  \centering
\includegraphics[width=\textwidth]{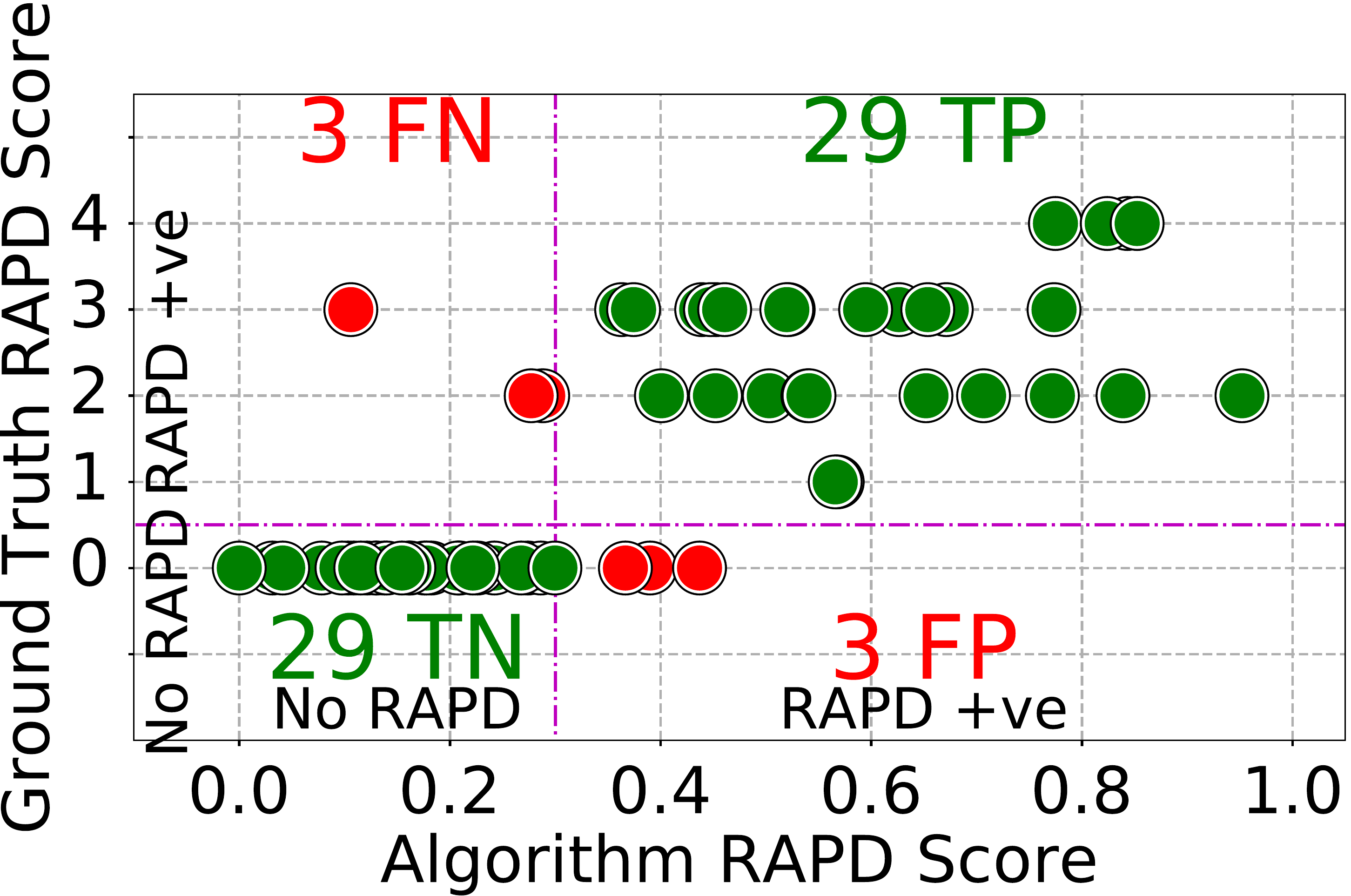}
  \vspace{0.01cm}
  \centerline{\footnotesize{(d) RAPDNet Scatter}}%\medskip
  \vspace{0.03 cm}
\end{minipage}
\centering
\vspace{-2 mm}
\caption{Scatter plots of RAPD detection algorithms.}
\label{fig:scatters}
\vspace{-4mm}
\end{figure}

We investigate the importance of \texttt{RAPD index} by performing a ROC curve analysis with alternative dissimilarity metrics in Fig.~\ref{fig:AUCs}. We measure the dissimilarity with 1 minus correlation metrics including Pearson, Spearman, and Kendall. In the ROC plot, each curve represents a pupil localization algorithm combined with a RAPD assessment metric. Y axis corresponds to true positive rate and x axis corresponds to false positive rate. We obtained the curves for each algorithm configuration by sweeping the detection threshold and highlighted the highest sensitivity and specificity with a colored circle. We can observe that \texttt{RAPD index} outperforms all the correlation metrics analyzed in this study. To further analyze the RAPD detection performance, we show the scatter plots of algorithmic configurations with highest detection performance for each localization algorithm in Fig.~\ref{fig:scatters}. Y axis separates the ground truth results as no RAPD and RAPD positive. X axis corresponds to scores obtained from RAPD detection algorithms. Overall, \texttt{RAPDNet} outperforms all other algorithms by maximum true decisions (29TP and 29TN) and minimum false decisions (3FN and 3FP).

Handcrafted algorithms Starburst, ExCuSe, and ElSe have numerous parameters that require tuning to obtain state-of-the-art performances, which makes it difficult to use these models in new datasets. However, in case of data-driven approaches, visual representations are learned from generic images rather than being handcrafted for specific tasks. Therefore, it can be relatively easier to use these generic representations for new tasks and datasets. In this study, we automatically learned the mapping between generic visual representations and target applications via transfer learning, which enabled achieving higher performance in the \texttt{RAPD-HD} dataset compared to benchmarked handcrafted algorithms.

\section{Conclusion}
\label{sec:conc}
    We introduced an algorithmic framework based on pupil localization and pupil size measurement to assess relative afferent pupillary defect. Experimental results show that state-of-the-art pupil localization algorithms can lead to an AUC of 0.848 over 64 test cases. In addition to benchmarking handcrafted state-of-the-art algorithms, we introduced a transfer learning-based RAPD detection algorithm that can achieve an AUC of 0.929, which outperformed all benchmarked algorithms. In this study, temporal information was used in the postprocessing stage but not directly in the pupil detection stage. However, temporal relationship between frames can be utilized to obtain a more robust pupil detection algorithm. Therefore, instead of utilizing spatial and temporal information independently, we need to investigate the utilization of spatiotemporal information available in the video sequences for pupillary light reflex assessment.  

%\bibliographystyle{IEEEbib}
%\bibliography{ref}

\begin{IEEEbiography}[{\includegraphics[width=1in,height=1.2in,clip,keepaspectratio]{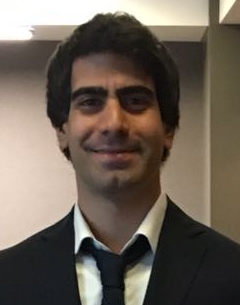}}]{Dogancan Temel} (M'13) received the M.S. and the Ph.D. degrees in electrical and computer engineering (ECE) from the Georgia Institute of Technology (Georgia Tech), Atlanta, USA, in 2013 and 2016, respectively. During the PhD program, he received the Texas Instruments Leadership Universities Fellowship for four consecutive years. Currently, he is a postdoctoral fellow at Georgia Tech focusing on human and machine vision. Dr. Temel was the recipient of the Best Doctoral Dissertation Award from the Sigma Xi honor society, the Graduate Research Assistant Excellence Award from the School of ECE, and the Outstanding Research Award from the Center for Signal and Information Processing at Georgia Tech. 
\end{IEEEbiography}
\vspace{-10 mm}
%\vskip -2.5\baselineskip plus -1fil
\begin{IEEEbiography}[{\includegraphics[width=1in,height=1.2in,clip,keepaspectratio]{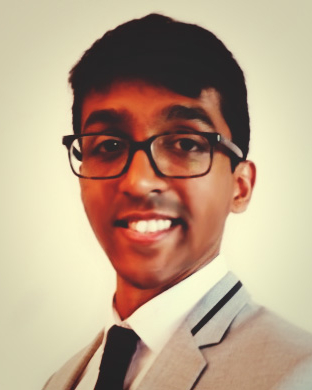}}]{Melvin J. Mathew} (M'18) received the B.Sc. degree in engineering in mechatronics, with honors, from the University of Cape Town (UCT), Cape Town, South Africa, in 2015. He also received the M.S. degree in electrical and computer engineering (ECE) from the Georgia Institute of Technology (Georgia Tech), Atlanta, USA, in 2018. Currently, he is a research engineer at the OLIVES lab at Georgia Tech, working on automated health-care and artificial intelligence. Mr. Mathew participated in the IEEE UBTECH-Education Robotics Design Challenge, held in China, 2017, placing 2nd with his team at Georgia Tech-Shenzhen, out of 100 teams involving 800 students.
\end{IEEEbiography}
\vspace{-10 mm}
%\vskip -2.5\baselineskip plus -1fil
\begin{IEEEbiography}[{\includegraphics[width=1in,height=1.2in,clip,keepaspectratio]{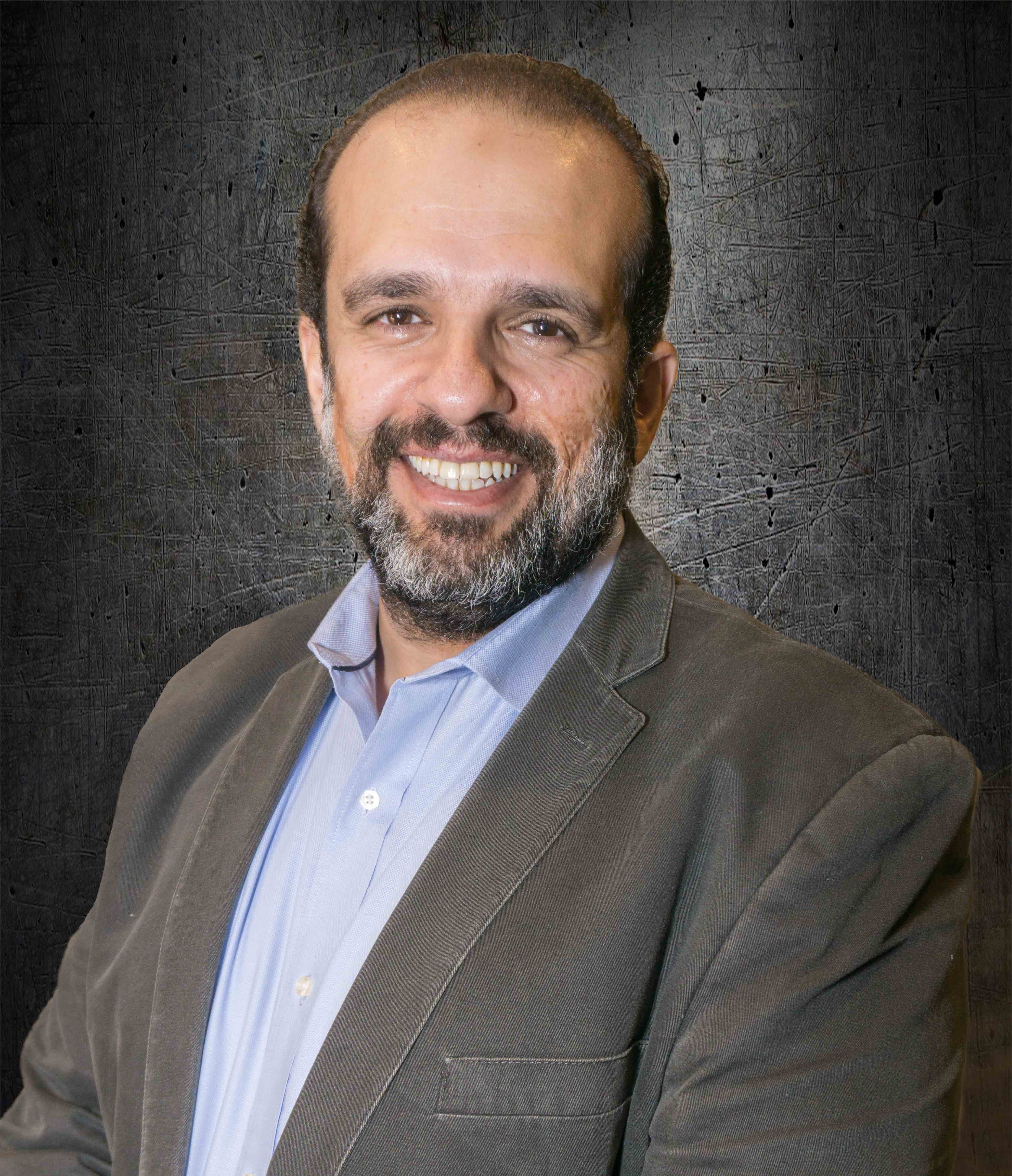}}]{Ghassan AlRegib} (M'97, SM'10) is currently a Professor with the School of Electrical and Computer Engineering, Georgia Institute of Technology.  He was a recipient of the ECE Outstanding Graduate Teaching Award in 2001 and both the CSIP Research and the CSIP Service Awards in 2003, the ECE Outstanding Junior Faculty Member Award, in 2008, and the 2017 Denning Faculty Award for Global Engagement.
His research group, the Omni Lab for Intelligent Visual Engineering  and Science (OLIVES) works on research projects related to machine learning, image and  video processing, image and video understanding, seismic interpretation, healthcare intelligence, machine learning for ophthalmology, and video analytics. He participated in several service activities within the IEEE including the organization of the First IEEE VIP Cup (2017), Area Editor for the IEEE Signal Processing Magazine, and the Technical Program Chair of GlobalSIP’14 and ICIP’20. He has provided services and consultation to several firms, companies, and international educational and Research and Development organizations. He has been a witness expert in a number of patents infringement cases.

 \\
\end{IEEEbiography}
\vspace{-10 mm}
%\vskip -2.5\baselineskip plus -1fil
\begin{IEEEbiography}[{\includegraphics[width=1in,height=1.2in,clip,keepaspectratio]{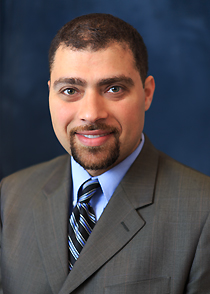}}]{Yousuf M. Khalifa} received the MD degree at the Medical College of Georgia, where he also completed an internship and ophthalmology residency. He then went to the University of California San Francisco’s Proctor Fellowship as a Heed Fellow, followed by a fellowship in cornea and refractive surgery at the University of Utah’s Moran Eye Center. Afterwards, he was cornea and external disease specialist and residency program director at the University of Rochester. Currently, he serves as chief of service of ophthalmology at Grady Memorial Hospital and in the cornea service at Emory Eye Center.   
%\enlargethispage{-1.0cm}
\end{IEEEbiography}

%{
%\renewcommand{\clearpage}{} 
%\renewcommand{\newpage}{} 
%\bibliography{ref}
%}

\end{document}